\documentclass{article} 

\usepackage{PRIMEarxiv}    

\usepackage[utf8]{inputenc} 
\usepackage[T1]{fontenc}    
\usepackage{hyperref}       
\usepackage{url}            
\usepackage{booktabs}       
\usepackage{amsfonts}       
\usepackage{nicefrac}       
\usepackage{microtype}      
\usepackage{lipsum}
\usepackage{fancyhdr}       
\usepackage{graphicx}       
\usepackage{amsmath}
\usepackage{amssymb}  
\usepackage{bm}
\usepackage{booktabs} 
\usepackage{natbib} 
\usepackage{tikz}
\usepackage{makecell}
\usepackage{xcolor}
\usetikzlibrary{positioning, arrows.meta}
\graphicspath{{media/}}     
\usepackage{xspace}
\usepackage{authblk} 
\newcommand{\lfm}{NASA-IBM LFM\xspace}

\usepackage{subcaption}

\title{Multimodal-Multiresolution Foundation Model for Lunar Remote Sensing}

\author[2]{Paolo Fraccaro$^\dagger$}
\author[2]{Gabby Nyirjesy$^\dagger$}
\author[2]{Daniela Szwarcman$^\dagger$}
\author[1]{Himanshu Patil$^\dagger$}
\author[1]{Vishal Gaur$^\dagger$}
\author[1]{Rohit Lal$^\dagger$}
\author[3]{Rachel A. Slank$^\dagger$}
\author[2]{Geoffrey Dawson$^\dagger$}
\author[2]{Hiyam Debary$^\dagger$}
\author[4,*]{Michael K. Barker$^\dagger$}
\author[5,6]{Andrew Annex$^\dagger$}
\author[4,7]{Vishnu Viswanathan}
\author[4,8]{Zachary Morse}
\author[5]{Ethan I. Schaefer}
\author[1]{Nikolaos Dionelis}
\author[1]{Ankur Kumar}
\author[2]{Campbell D. Watson}
\author[10]{Manil Maskey}
\author[9]{Rebekah I. Dawson-Rigas}
\author[2]{Juan Bernab\'e-Moreno}
\author[10]{Rahul Ramachandran}
\author[1,10,*]{Sujit Roy}

\affil[1]{The University of Alabama in Huntsville, Huntsville, AL, USA}
\affil[2]{IBM Research, Yorktown Heights, NY, USA}
\affil[3]{University Space Research Association (USRA), Science and Technology Institute, Huntsville, AL, USA}
\affil[4]{NASA Goddard Space Flight Center, Greenbelt, MD, USA}
\affil[5]{SETI Institute, Mountain View, CA, USA}
\affil[6]{NASA Ames Research Center, Moffett Field, CA, USA}
\affil[7]{University of Maryland, Baltimore County (UMBC), Baltimore, MD, USA}
\affil[8]{Howard University, Washington, DC, USA}
\affil[9]{NASA Headquarters, Washington, DC, USA}
\affil[10]{NASA Marshall Space Flight Center, Huntsville, AL, USA}

\affil[*]{Corresponding authors: Sujit Roy (sujit.roy@nasa.gov) and Paolo Fraccaro (paolo.fraccaro@ibm.com)}

\affil[$\dagger$]{Equal contribution to this work}

\begin{document}          

\maketitle              

\begin{abstract}

We present a multimodal foundation model for lunar remote sensing, pretrained from scratch on SomBench, a co-registered corpus of nearly two million tile bundles spanning 11 modalities at two spatial scales (i.e., 1 m/pixel and 100 m/pixel). The dataset is geographically partitioned to prevent leakage across train, validation, and test splits. The model adapts the TerraMind masked-token architecture to the lunar setting with two extensions: per-tile acquisition geometry (i.e., illumination angles, solar-frame anchors, tile footprint) is provided as explicit contextual information rather than requiring the model to infer it implicitly from image content; and tiles anchored at meter and hundred-meter scale are trained jointly in a mixed-batch regime so that a single set of weights covers both resolution families. A FlexiViT patch-embedding scheme allows the same pretrained checkpoint to be reused at different working patch sizes without backbone retraining, and modality-wise tokenization lets downstream users drop or add modalities at fine-tuning time. Qualitative generation experiments indicate that the model captures meaningful cross-modal correspondences (i.e., terrain derivatives from elevation, illumination-consistent reflectance from geometry, and coarse spatial pattern across co-registered modality sets). Evaluated across four downstream benchmarks (i.e., crater detection at 100 m/pixel and NAC 1 m/pixel imagery, irregular mare patch segmentation, and polar ice prospectivity regression) using TerraTorch, the pretrained model matches or outperforms ImageNet-pretrained baselines and an architecturally identical random-initialization control. On multimodal ice prospectivity regression at the poles, the pretrained model variants (i.e., full finetuning, Low-Rank Adaptation (LoRA) and frozen) outperform all baselines by margins that exceed run-to-run variability, with the randomly initialized version of our model that beats all but one baselines. This indicates that part of the gain comes from our architecture's native token-level multimodal processing rather than from pretraining alone. Label efficiency is particularly pronounced for crater detection: at 50\% training data on crater detection at 100 m/pixel, the pretrained model already exceeds the strongest ImageNet baseline trained on the full dataset. Among adaptation strategies, LoRA matches or surpasses full fine-tuning on crater detection and irregular mare patches segmentation, using a small fraction of the trainable parameters with lower run-to-run variance. Full fine-tuning remains best on ice-prospectivity regression, with LoRA that however still beats all baselines. We release the pretrained checkpoint, fine-tuning code, and benchmark datasets to support reproducibility and community reuse.

\end{abstract}


\section{Introduction}

The lunar surface is observed through a heterogeneous collection of orbital instruments rather than a single canonical image stream. The Lunar Reconnaissance Orbiter Camera (LROC) provides wide-angle Wide Angle Camera (WAC) and meter-scale Narrow Angle Camera (NAC) imagery \citep{Robinson2010Lunar,Robinson2010Lunara}; the Lunar Orbiter Laser Altimeter (LOLA) measures topography \citep{Smith2010Lunar}; Diviner Lunar Radiometer Experiment (Diviner) records reflected and emitted radiation across the thermal spectrum \citep{Paige2009Lunar}; Mini-RF supplies radar observations \citep{Nozette2010Lunar,Fassett2024Improved}; the Kaguya/SELENE Multiband Imager characterizes visible and near-infrared mineralogy \citep{Ohtake2008Performance}; Lunar Prospector Neutron Spectrometer provides hydrogen abundance maps for the lunar poles \cite{feldman1999lunar,Lawrence2022}; and GRAIL (Gravity Recovery and Interior Laboratory) maps variations in the lunar gravity field \citep{Zuber2013Gravity}. These measurements encode complementary aspects of morphology, composition, thermophysics, and subsurface structure, but their native spatial resolutions span approximately four orders of magnitude. Terrain products alone range from near-global 100~m WAC stereo topography \citep{Scholten2012GLD} and the approximately 60~m SLDEM2015 (SELENE Digital Elevation Model) product \citep{Barker2016new} to meter-scale NAC stereo elevation models \citep{Burns2012DIGITAL,Henriksen2016Extracting}. Consequently, learning a reusable lunar representation is not simply a matter of scaling natural-image pretraining: it requires a model that can align different physical variables, tolerate absent observations, and preserve useful structure while moving between regional and local spatial scales.

Masked autoencoding offers an attractive basis for such a model because it converts abundant unlabeled imagery into a scalable reconstruction objective \citep{he2022masked}, and MultiMAE extends this principle across modalities \citep{Bachmann2022MultiMAE}. Nevertheless, existing geospatial adaptations \citep{Cong2022SatMAE,Fuller2023CROMA,Jakubik2023Foundation} transfer imperfectly to the Moon. First, raw-pixel reconstruction targets are noisy, spending model capacity on high-frequency detail that carries little semantic content --- a poor trade on low-signal, illumination-dominated lunar imagery. Apparent surface appearance on the Moon is governed more by acquisition and illumination geometry than by intrinsic surface variation, so a model left to disentangle illumination from pixels alone spends capacity recovering a quantity that is already recorded for every tile. Second, because native resolutions span four orders of magnitude, a recipe that trains one resolution family at a time cannot learn how regional context and meter-scale detail constrain one another.

We address these issues with the NASA-IBM Lunar Foundation Model (\lfm), pretrained from scratch on SomBench \cite{sombench2026collection}: nearly two million co-registered tile bundles organized as two data tracks. The WAC-anchored track covers optical, ultraviolet, terrain, slope and aspect; the NAC-anchored track pairs panchromatic meter-scale imagery with co-registered 3~m terrain, slope, and aspect. Both tracks also include contextual information for illumination geometry, thermal, radar, mineralogical, gravity, and various geologic products. Tiles are partitioned by Lunar Transverse Mercator (LTM) zones rather than independently, so geographically overlapping observations cannot straddle the training, validation, and test splits (Section~\ref{sec:dataset}). The corpus is deliberately broad enough to support both regional reasoning and meter-scale analysis of features such as small craters and irregular mare patches, whose scientific catalogs illustrate the breadth and difficulty of lunar mapping \citep{Robbins2019New,Qiao2020Lunar,hargitai2025clusters}. 

Our starting point is TerraMind \citep{jakubik2025terramind}, a best-in-class multimodal recipe developed for Earth observation, which we retrain from scratch rather than fine-tune from its Earth-observation checkpoint. Two lunar-specific extensions follow directly from the second and third problems above. Per-tile acquisition geometry --- illumination angles, solar-frame anchors, and tile footprint --- is promoted to its own sequence-tokenized encoder input, so the dominant confound on lunar appearance is handed to the model rather than inferred by it. Furthermore, the two anchored tracks are trained jointly at their native resolutions in a single mixed-batch loop, so one set of weights covers both scales instead of one optimization stream per resolution family. Modality-wise tokenization and FlexiViT patch-embedding resizing \citep{beyer2023flexivit} then make the resulting checkpoint reusable under input conditions it never saw during pretraining. Section~\ref{sec:method} gives the full architecture and pretraining recipe.

Our contributions are:
\begin{itemize}
\item \textbf{Geometry-conditioned, mixed-resolution pretraining.} We show that supplying acquisition geometry as encoder tokens and training both resolution families in one loop are compatible with a standard multimodal masked-token objective at scale.
\item \textbf{A flexible checkpoint for downstream task adaptation.} The same weights adapt to patch grids other than the pretraining one, accept modality subsets that differ from the pretraining mixture, and support Low-Rank Adaptation (LoRA) applied to the pretrained encoder.
\item \textbf{Evidence of importance of lunar pretraining} Across four benchmarks (i.e., crater detection at WAC and NAC scale, irregular mare patch segmentation, and polar ice prospectivity regression, all run through a common fine-tuning framework \citep{Gomes2025TerraTorch}) the pretrained model matches or exceeds ImageNet-pretrained baselines and an architecture-matched random-init control.
\end{itemize}

The remainder of the paper is organized as follows. Section~\ref{sec:related} compares the model against masked image modeling, available lunar foundation models, and prior machine learning on lunar observations. Section~\ref{sec:dataset} describes the pretraining corpus and the four benchmarks. Section~\ref{sec:method} presents the architecture, the pretraining recipe, and the fine-tuning protocol. Section~\ref{sec:experiments} first probes what pretraining has learned through multimodal generation, then reports transfer performance against ImageNet-pretrained baselines and a random-initialization control.

\section{Related Work}
\label{sec:related}

\paragraph{Masked image modeling and reconstruction targets.}
Masked autoencoders learn visual representations by reconstructing randomly hidden patches with an asymmetric encoder--decoder architecture \citep{he2022masked}. MultiMAE generalizes this objective to multiple input and output domains, sampling a limited visible-token budget across modalities to encourage cross-modal prediction \citep{Bachmann2022MultiMAE}.


Discrete-target formulations replace pixels with tokenized codes: BEiT reconstructs discrete visual codes from a pretrained tokenizer \citep{Bao2022BEiT}; MaskFeat reconstructs handcrafted local descriptors \citep{Wei2022Masked}; and Feature-guided MAE augments pixel targets with spectral and spatial descriptors to counteract the pixel-fidelity bias in noisy SAR imagery \citep{Wang2024Feature}, which is a bias that also afflicts low-signal lunar imagery. TerraMind, inspired by 4M \cite{mizrahi20234m}, sits in this lineage, using a VQ-VAE tokenizers to produce per-modality discrete codes as the reconstruction target and predicting a categorical distribution over the resulting vocabulary at masked positions \citep{jakubik2025terramind}. We adopt this tokenization scheme unchanged.

\paragraph{Lunar foundation models.}
Compared to the rich literature on Earth observation foundation models, prior lunar foundation-model efforts remain limited. LunarFM \citep{girona2026lunarfm} is trained on curated global WAC-resolution lunar products, emphasizing multimodal representation learning at regional and global scales. A TerraMind-style unified transformer for lunar NAC imagery \citep{sander2026moon} restricts pretraining to high-resolution generation and shape-from-shading on a handful of Apollo-site NAC scenes, targeting monocular DEM (Digital Elevation Model) prediction rather than a reusable multimodal representation. Our model differs from these because it brings together three aspects: pretraining is globally distributed rather than site-limited, spans both resolution families rather than one, and targets a reusable representation rather than a single reconstruction task or a curated map snapshot.


\paragraph{Lunar observations and machine learning.}

Impact-crater mapping is a central downstream application. The Robbins catalog contains more than two million lunar craters and provides a global reference for crater location and size \citep{Robbins2019New}. Prior learning-based methods include residual U-Net detection from lunar DEMs \citep{Wang2020Effective} and a super-resolution-assisted YOLO architecture aimed at improving small-crater detection under limited resolution and varying illumination \citep{Grassa2023YOLOLens}. We evaluate crater transfer at both WAC and NAC scales. Irregular mare patches (IMPs) form a distinct fine-boundary segmentation problem. Their complex morphologies and debated origin have motivated an updated catalog and geologic classification \citep{hargitai2025clusters}, making them a useful test of whether meter-scale pretraining captures more than crater morphology.

\paragraph{Terrain reconstruction and geometric consistency.}
High-resolution lunar terrain has traditionally been obtained from stereo observations or shape-from-shading. Monocular methods combine image shading with a coarse DEM constraint to produce pixel-scale elevation \citep{Wu2017Construction}. Learning-based approaches predict large-area NAC topography constrained by SLDEM2015 \citep{Chen2022CNN}, combine a CNN estimate with shape-from-shading refinement \citep{Chen2022PIXEL}, or use adversarial learning to generate high-resolution DEMs from monocular images and low-resolution terrain \citep{Liu2022Generative}. ELunarDTMNet introduces hierarchical multiscale modeling for single-view lunar DTM (Digital Terrain Model) reconstruction and evaluates cross-dataset generalization \citep{Chen2024ELunarDTMNet}. Beyond the Moon, DEM super-resolution methods improve local and global terrain features through feature enhancement, residual fusion, attention, and topography-aware transformers \citep{Ma2023Feature,Chen2023Enhanced,Wang2024TTSR}. In our model, DTM, slope, and aspect enter the pretraining objective as three separate co-ingested modalities, each with its own tokenizer and its own masked-token reconstruction target, and the shared encoder is left to learn their inter-relationship from the co-registered pretraining data rather than through an explicit differential constraint.

\section{Datasets}
\label{sec:dataset}
To train and evaluate \lfm we use the pretraining and benchmarking datasets from SomBench \cite{sombench2026collection}, which we built and released as a companion dataset publication. Full details are in the original paper; we summarize the corpus here for the reader's convenience.

\subsection{Pretraining dataset}

\paragraph{Motivation}
\label{sec:dataset_motivation}

Lunar remote-sensing archives are large but instrument-siloed: each product ships in its own projection, resolution, and coverage pattern, and most lunar science tasks remain label-poor. We therefore constructed, to our knowledge, the largest co-registered multimodal lunar corpus: approximately two million multimodal tile bundles spanning more than thirty lunar data products at two spatial scales. Its design targets three capabilities that no existing lunar dataset provides jointly. First, pixel-aligned modality bundles make cross-modal structure learnable from spatially corresponding observations and support any-to-any modality completion. Second, two spatial regimes enable multi-resolution representation learning across a 100$\times$ scale gap. These pivot around regional geology at Wide Angle Camera (WAC) scale and meter-scale geomorphology at Narrow Angle Camera (NAC) scale, both with geometry acquisition metadata. Third, Geographic partitioning makes downstream evaluation spatially leakage-free.

\paragraph{Data Sources}
\label{sec:dataset_sources}

The optical anchors are fully calibrated and map projected Lunar Reconnaissance Orbiter Camera (LROC) experiment data records (EDRs). The low-resolution track is anchored to 54{,}080 WAC observation pairs, each providing five-band visible reflectance (415--689\,nm) at 100~m/px and two-band ultraviolet reflectance (321--360\,nm) at 500~m/px. WAC was chosen as the regional anchor because it offers near-global coverage at a resolution that preserves regional geological context, with overlapping observations exposing the same terrain under different illumination. The high-resolution track is anchored to 1{,}095 NAC frames at $\sim$1~m/px, restricted to the NAC PHO (photometry) sites and the orthorectified frames used to produce the LRO 3~m stereo Digital Terrain Models (DTMs). This ensures that every NAC tile is paired with a co-registered 3~m terrain model rather than 60~m topography alone.

Topography enters at both scales: the LOLA--Kaguya merged SLDEM2015 model (60~m) for the WAC track and NAC-stereo DTMs (3~m) for the NAC one. For each elevation model slope and aspect are also computed at their native resolution, with aspect stored as a sine--cosine pair to avoid the angular wrap-around. This sine--cosine representation is what we will refer to as \textit{aspect} throughout the remainder of the paper. As a photometrically controlled reflectance layer we use the WAC 643\,nm normalized-reflectance product (100~m), which is available for every WAC anchor. Beyond these, the corpus attaches $\sim$20 further co-registered data sources spanning across geologic mapping, Diviner thermophysics, LOLA products (roughness, permanently shadowed regions, illumination, 1064\,nm normal albedo), Mini-RF radar, Kaguya MI/SP mineralogy and space weathering, GRAIL gravity disturbance, and Lunar Prospector hydrogen abundance (Table~\ref{tab:flfm-modalities}). Every tile also carries per-product acquisition metadata (solar incidence, emission, phase, and azimuth angles, sub-solar and tile-center coordinates, and ground sampling distance) joined from the LROC EDR geodatabase; these fields are available at inference whenever the image itself is, so they introduce no label leakage.

\paragraph{Dataset Construction Pipeline}
\label{sec:dataset_pipeline}

\begin{figure}[t]
\centering
\begin{tikzpicture}[
  node distance=3mm,
  stage/.style={draw, rounded corners=2pt, align=center, fill=gray!8,
                inner sep=4pt, font=\footnotesize, text width=52mm},
  arr/.style={-{Stealth[length=2mm]}, semithick}
]
\node[stage] (raw)   {\textbf{Raw products}\\
  \scriptsize LROC WAC/NAC EDRs $+$ $\sim$30 map layers};
\node[stage, below=of raw] (anchor) {\textbf{Optical anchors}\\
  \scriptsize WAC 100~m/px $\cdot$ NAC 1~m/px};
\node[stage, below=of anchor] (proj) {\textbf{Equal-area projection}\\
  \scriptsize per-strip LTM zones $\cdot$ per-pole LPS caps};
\node[stage, below=of proj] (coreg) {\textbf{Co-registration}\\
  \scriptsize layers snapped to the anchor pixel grid};
\node[stage, below=of coreg] (tile) {\textbf{Tile extraction}\\
  \scriptsize 512$\times$512\,px sliding window, zone-tagged};
\node[stage, below=of tile] (bundle) {\textbf{Multimodal bundles}\\
  \scriptsize per-modality netCDF $\to$ index-aligned shards};
\node[stage, below=of bundle] (split) {\textbf{Spatial split}\\
  \scriptsize train/val/test by LTM zone and LPS partition};
\draw[arr] (raw) -- (anchor);
\draw[arr] (anchor) -- (proj);
\draw[arr] (proj) -- (coreg);
\draw[arr] (coreg) -- (tile);
\draw[arr] (tile) -- (bundle);
\draw[arr] (bundle) -- (split);
\end{tikzpicture}
\caption{Dataset construction. Each optical observation defines the sample extent; auxiliary products are co-registered onto the anchor grid, extracted as multimodal tile bundles, and partitioned geographically.}
\label{fig:dataset-pipeline}
\end{figure}
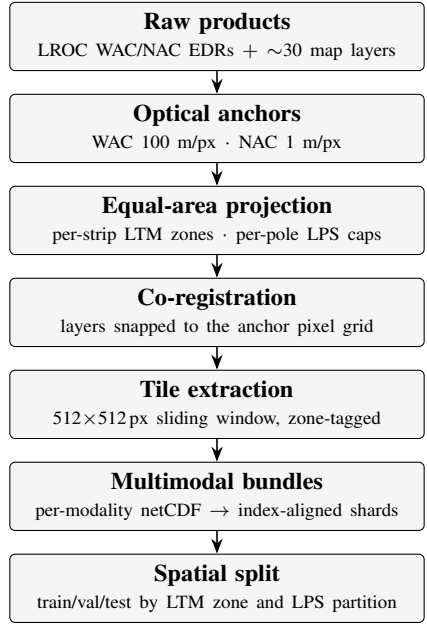

Tiles are generated by an image-anchored pipeline in which each optical observation defines the spatial extent of its samples (Figure~\ref{fig:dataset-pipeline}): a sliding window over the anchor canvas creates 512$\times$512\,px tiles, and all auxiliary products are attached at each tile's exact geographic bounds. Construction operates in equal-area map projections---per-strip Lunar Transverse Mercator (LTM) zones and per-pole Lunar Polar Stereographic (LPS) caps---with every source product pre-warped once per zone, so tile extraction never re-projects the anchor imagery itself. Each tile is tagged with its zone, and tiles that straddle zone boundaries are dropped, which later guarantees clean split boundaries (Section~\ref{sec:dataset_splits}).

Auxiliary data sources are re-gridded onto a per-channel resolution snapped to the anchor grid, so data sources of a bundle share corner pixels despite native resolutions spanning four orders of magnitude. Continuous modalities are bilinearly resampled, categorical products (e.g., the geologic map) use nearest-neighbor interpolation, and multi-band products are stacked along a band dimension. Missing observations are preserved explicitly rather than imputed: data exceeding a missing-pixel threshold are recorded as absent. Bundles are stored as compressed per-modality numpy files and packaged into index-aligned WebDataset shards for high-throughput multimodal loading. 

\paragraph{Spatial Splits}
\label{sec:dataset_splits}

Random tile-level splits are not suitable for our dataset. The sliding window creates overlapping tiles, and repeated EDR coverage observes the same ground patch in many tile bundles, so a random partition would place near-duplicates on both sides of the evaluation boundary. The corpus is therefore split geographically. Every tile carries its LTM zone or LPS cap; whole zones are assigned to train, validation, or test ($\approx$80/10/10), the polar caps are spatially partitioned with the same guarantee, and tiles that straddle zone boundaries were already dropped at construction time. Zone assignments are shared between the WAC and NAC tracks, so a ground patch admitted to both views is routed to the same split at both scales. Train and validation drive pretraining and checkpoint selection; the test partition is reserved for downstream fine-tuning and evaluation (Section~\ref{sec:experiments}).

\paragraph{Statistics}
\label{sec:dataset_stats}

Table~\ref{tab:flfm-modalities} summarizes the corpus composition. The final corpus contains 963{,}609 WAC bundles and 1{,}000{,}113 NAC bundles, built from 54{,}080 WAC pairs and 1{,}095 NAC frames and totalling 43 million modality files ($\sim$53\,TB compressed; 51.4\,TB WAC, 1.6\,TB NAC). All pretraining modalities are present for essentially every tile of their track (null fractions $\leq$0.3\%), while coverage of the wider corpus varies by instrument footprint---from near-global layers (geologic map, $T_{\mathrm{bol}}$, gravity) through mid-latitude products (Mini-RF, Kaguya MI) to polar-only layers (PSR, average illumination, ice-stability depth), which exist for $\sim$6{,}900 polar tiles per track.

\begin{table}[t]
\centering
\caption{Corpus composition. Channels marked ``multi'' are multi-band stacks (e.g., 24 Diviner $T_{\mathrm{bol}}$ phases plus the acquisition-matched phase). The \textbf{Pretrain} column gives the representation each layer receives during pretraining: \emph{dense} layers are patch-embedded and tokenized as full images, forming the nine image-like modalities; \emph{scalar} layers are averaged over the tile footprint and contribute a single discrete token each, together forming the static-map context modality. With the per-tile optical metadata, this gives the 11 pretraining modalities.}
\label{tab:flfm-modalities}
\footnotesize
\setlength{\tabcolsep}{3pt}
\begin{tabular}{@{}llrcl c@{}}
\toprule
\textbf{Modality} & \textbf{Key} & \textbf{Native res.} & \textbf{Ch.} &
\textbf{Source} & \textbf{Pretrain} \\
\midrule
\multicolumn{6}{@{}l}{\emph{WAC family (low-resolution track, 51.2\,km tiles)}} \\
Visible reflectance & \texttt{vis} & 100~m & 5 & LROC WAC & Dense \\
Ultraviolet reflectance & \texttt{uv} & 500~m & 2 & LROC WAC & Dense \\
Topography & \texttt{dtm} & 60~m & 1 & SLDEM2015 & Dense \\
Slope & \texttt{slope} & 60~m & 1 & SLDEM2015 & Dense \\
Aspect & \texttt{aspect} & 60~m & 2 & SLDEM2015 & Dense \\
\midrule
\multicolumn{6}{@{}l}{\emph{NAC family (high-resolution track, 512~m tiles)}} \\
Panchromatic imagery & \texttt{nac} & 1~m & 1 & LROC NAC & Dense \\
Topography & \texttt{dtm\_3m} & 3~m & 1 & NAC-stereo DTM & Dense \\
Slope & \texttt{slope\_3m} & 3~m & 1 & NAC-stereo DTM & Dense \\
Aspect & \texttt{aspect\_3m} & 3~m & 2 & NAC-stereo DTM & Dense \\
\midrule
\multicolumn{6}{@{}l}{\emph{Additional corpus layers (attached to both tracks as a tile-level average, where available)}} \\
Norm.\ 643\,nm reflectance & \texttt{nr643} & 100~m & 1 & LROC WAC & Scalar \\
Regolith temp.\ anomaly & \texttt{TREG} & 240~m & 1 & Diviner & Scalar \\
Bolometric temp.\ (phases) & \texttt{TBOL} & 15\,km & multi & Diviner & Scalar \\
Rock abundance & \texttt{ROCK\_ABUND.} & 240~m & 1 & Diviner & Scalar \\
H-parameter & \texttt{HPAR} & 240~m & 1 & Diviner & Scalar \\
Ice-stability depth & \texttt{DICE} & 240~m & 1 & Diviner & Scalar \\
Roughness & \texttt{ROUGHNESS} & 1\,km & 1 & LOLA & Scalar \\
PSR / illumination & \texttt{PSR}, \texttt{AVG\_ILLUM} & 20--120~m & 1 & LOLA & Scalar \\
1064\,nm albedo & \texttt{ALBEDO} & 1\,km & 1 & LOLA & Scalar \\
Radar CPR / reflectivity & \texttt{MINIRF\_*} & 90~m & 1 & Mini-RF & Scalar \\
MI mineralogy / SW-Fe & \texttt{MI\_*}, \texttt{SW\_FE} & 60~m--1\,km & multi & Kaguya & Scalar \\
Polar mineralogy & \texttt{SP\_MINER.} & 1\,km & multi & Kaguya SP & Scalar \\
TiO$_2$ / morph mosaic / refl. & \texttt{TIO2}, \texttt{WAC\_*} & 100--500~m & 1 & LROC WAC & Scalar \\
Gravity disturbance & \texttt{GRAVITY} & 20\,km & 1 & GRAIL & Scalar \\
Hydrogen abundance & \texttt{HYDROGEN} & 15\,km & 1 & Lunar Prospector & Scalar \\
\bottomrule
\end{tabular}
\end{table}

\subsection{Benchmarking datasets}

Table \ref{tab:downstream-bench-overview} shows a summary of the SomBench benchmark datasets we used to validate our model. A description for each of the different task is provided below.

\begin{table*}[t]
\centering
\small
\caption{Overview of SomBench benchmark datasets used for the validation of our model.}
\label{tab:downstream-bench-overview}
\setlength{\tabcolsep}{4pt}
\renewcommand{\arraystretch}{1.05}
\resizebox{\textwidth}{!}{%
\begin{tabular}{@{}l l c c c r r r@{}}
\toprule
\textbf{Name} &
\textbf{Science Theme} &
\textbf{Image Size (px)} &
\textbf{Resolution (m/px)} &
\textbf{Target} &
\textbf{Train} &
\textbf{Val} &
\textbf{Test} \\
\midrule
\multicolumn{8}{@{}l}{\textbf{Object Detection}} \\
\midrule
Robbins craters & Impact processes & 512 $\times$ 512 & 100 & 1 class & 800 & 100 & 100 \\
NAC hand labeled & Impact processes & 256 $\times$ 256 &  1 & 1 class & 645 & 59 & 62 \\
\midrule
\multicolumn{8}{@{}l}{\textbf{Segmentation}} \\
\midrule
Irregular Mare Patches & Volcanic History & 256 $\times$ 256 & 1 & 1 class & 100 & 20 & 10 \\
\midrule
\multicolumn{8}{@{}l}{\textbf{Regression}} \\
\midrule
Lunar ice prospectivity & Polar Volatiles & 256 $\times$ 256 & 240 & 0--1 & 108 & 23 & 25 \\
\bottomrule
\end{tabular}
}
\end{table*}

\paragraph{Impact Processes}
\leavevmode\\
\leavevmode\\
Impact craters are the most ubiquitous lunar landform and span orders of magnitude in scale across all terrain types, making them a natural ML benchmark for detection, segmentation, and size--frequency analysis. Crater catalogs also underpin downstream applications from relative-age dating to geologic mapping and landing-site hazard assessment.

\subparagraph{Robbins Crater Catalog}

The Robbins catalog \cite{Robbins2019New} is a manually annotated global database of over 2 million lunar impact craters, intended as a near-complete census of diameters $\geq$1--2,km, digitized from LROC WAC global mosaics with topographic cross-checks under challenging illumination. The rim polygons are converted to per-tile bounding-box annotations. This is used to create a context-scale object-detection benchmark of 1,000 WAC visible tiles drawn from the test split of the pretraining dataset, filtered to solar-incidence angles between 60$^\circ$ and 80$^\circ$ for favorable illumination.

\subparagraph{NAC Hand Labeled datasets}

A meter-scale crater benchmark is built from LROC NAC imagery (1--5 m/pixel) spanning six NAC PHO sites (Highlands, King Ejecta, Apollo 15 S-IVB, Apollo 17, Reiner Gamma, and March 17 Impact Crater) selected under relief-enhancing illumination conditions (incidence angles of \textasciitilde{}50--80$^\circ$). Using co-registered 3 m/pixel DTMs alongside the NAC images, craters are manually labeled within 4--6 fixed 1024$\times$1024 px study areas per site and digitized as circles using OpenCraterTool \citep{Heyer2023}. The annotations are then mapped to 256$\times$256 NAC patches and packaged in COCO format, yielding 766 patches with 97,104 crater annotations. Dataset splits are enforced at both the site and study-area levels to prevent data leakage. Sites annotated at 5 m/pixel are visibly blurrier than those acquired at higher spatial resolution.

\paragraph{Volcanic History}

Irregular mare patches (IMPs) complement impact-feature benchmarks for assessing volcanic history. They are rare, morphologically diverse landforms that are typically distinguished by subtle boundary expressions rather than strong visual signatures. The benchmark was constructed starting from 2,623 polygon annotations \cite{hargitai2025clusters} paired with corresponding NAC images. In particular, a subset of annotations was manually adjusted to correct misalignment caused by NAC pointing uncertainty (typically on the order of tens of meters in uncontrolled frames), and approximately 100 polygons exceeding the 256 m patch size were excluded. These limitations highlight the potential value of future benchmark datasets derived from controlled NAC products, such as NAC PHO.

\paragraph{Polar Volatiles}

To support polar-volatiles benchmarking, we test \lfm at incorporating the polar map stack of \cite{coyan2025prospectivity}, packaged as eight static layers within 10$^\circ$ latitude of each pole at 240~m/pixel. The layers combine thermophysical constraints (ice stability depth, maximum surface temperature), illumination state (permanently shadowed regions (PSRs), computed via horizon/visibility modeling), terrain controls (slope, aspect, curvature), and two spatial-context features capturing cold-trap ``halo'' effects (distance to PSRs, PSR density). The regression target is the final continuous ice-prospectivity map that estimates where water ice is most likely within the upper $\sim$1~m of regolith \cite{coyan2025prospectivity}. This was originally calculated as knowledge-driven fuzzy overlay of the eight inputs \cite{coyan2025prospectivity}. Topographic derivatives are computed from the DTM and thermal layers from a numerical thermal model calibrated to Diviner observations. This compact, physically motivated stack enables standardized evaluation of polar terrain classification and ice-prospectivity prediction under extreme illumination regimes, with consistent spatial extent, resolution, and projection conventions.

\section{Method}
\label{sec:method}


\lfm's architecture is shown in Figure \ref{fig:fm-overview}. It follows the TerraMind \cite{jakubik2025terramind} design, which formulates multimodal learning as a token-prediction problem. Heterogeneous modalities are mapped into a shared embedding space and processed by an encoder-decoder Transformer. During pretraining, the model learns correlations across and within modalities by predicting sampled target tokens from sampled multimodal inputs.

\subsection{Modalities and representations}
\label{sec:modalities-representations}

As described in Section \ref{sec:dataset}, we pretrained \lfm on two families of multimodal samples centered on LROC NAC and WAC optical observations, respectively. Each training sample belongs to either a NAC-centered or a WAC-centered modality set and does not combine observations from the two. Within each modality set, corresponding regions from the available dense modalities describe \emph{the same geographic region}.

\begin{figure}[ht]
  \centering
  \includegraphics[width=\textwidth]{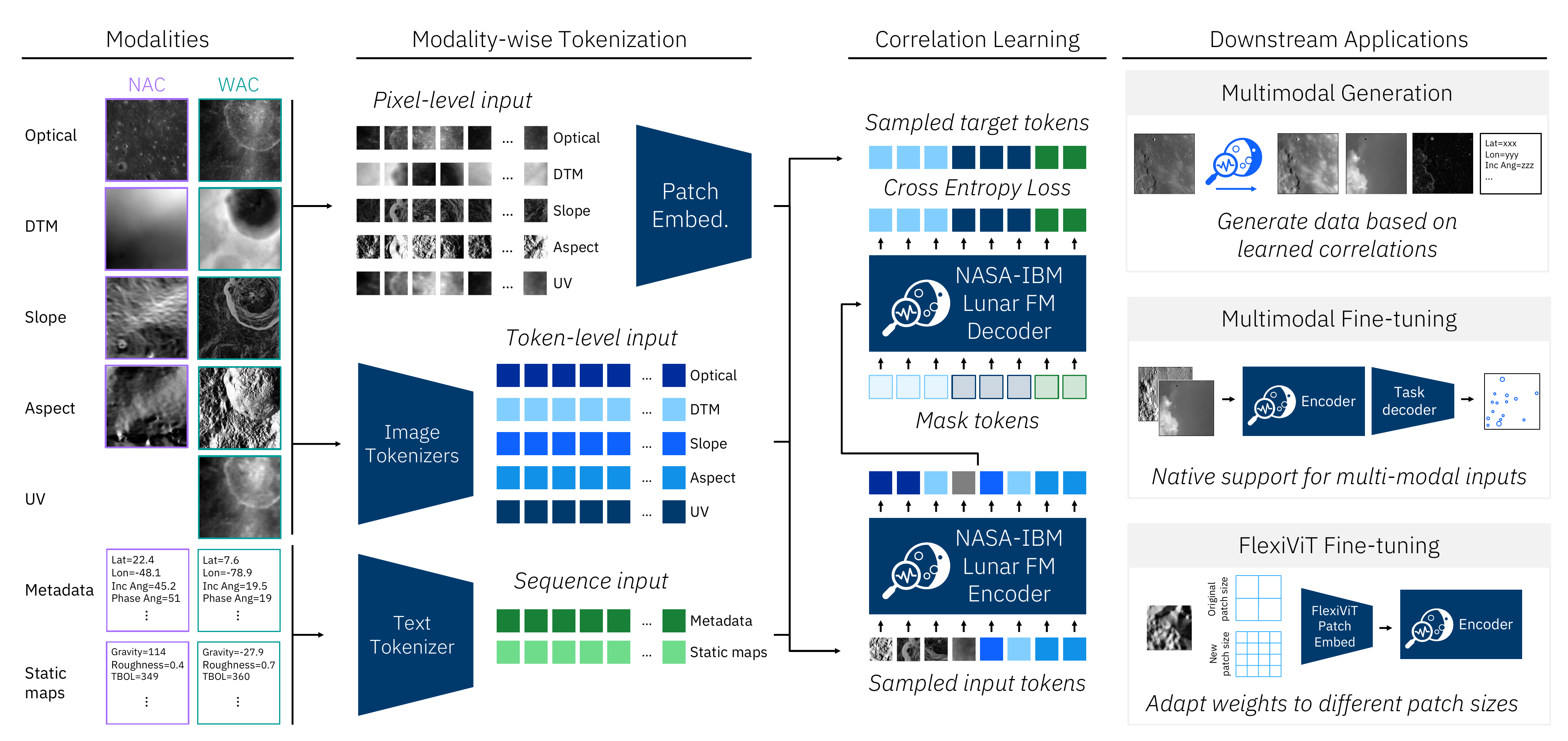}
    \caption{Model overview. \emph{Left:} input modalities include low- and high-resolution images (LROC WAC and NAC), DTM-derived topographic variables (DTM, slope, aspect), per-tile optical metadata, and static context maps. \emph{Center-left:} modality-wise tokenization in parallel branches. Image modalities are processed by patch-embedding layers (\emph{Pixel-level input}) and VQ-VAE tokenizers (\emph{Token-level input}); metadata and static maps are encoded with a text tokenizer (\emph{Sequence input}). \emph{Center-right:} correlation learning follow the TerraMind training procedure. A subset of the modality tokens is sampled as encoder inputs, while a separate subset is sampled as prediction targets. The encoder processes the input tokens, and the decoder predicts the target tokens under a cross-entropy objective. \emph{Right:} The pretrained model can be used for multimodal generation and multimodal fine-tuning with native support for multiple input modalities. The added FlexiViT support enables fine-tuning at patch sizes different from those used in pretraining.}
  \label{fig:fm-overview}
\end{figure}

The model processes dense image-like modalities and sequence-like contextual modalities. The dense modalities comprise the optical observation (WAC or NAC), DTM-derived topographic modalities (elevation, slope, and aspect), and UV (the latter being available only in the WAC-centered set). The sequence-like modalities comprise per-tile optical metadata and contextual variables derived from static maps. Table~\ref{tab:context-features} details the fields considered and the type of information they provide. The three input processing branches shown in Figure \ref{fig:fm-overview} (center-left) are:
\begin{itemize}
  \item \emph{Pixel-level image input}: Each dense modality is divided into 16$\times$16-pixel patches. Learnable linear projection layers map the pixel values of each patch to the shared encoder embedding dimension, providing the model with fine-grained representations of the dense modalities.
  \item \emph{Token-level image input}: The dense modalities are also processed by modality-specific VQ-VAE tokenizers trained on the pretraining corpus. Each tokenizer encodes a 16$\times$16-pixel region into a latent representation and quantizes it into a discrete code using Finite Scalar Quantization (FSQ)~\cite{fsq}. The resulting codes are mapped to the shared encoder embedding dimension through modality-specific learnable embedding layers.
  \item \emph{Sequence-level input}: The eight optical metadata variables and the 28 static-map-derived context variables are encoded as sequences of discrete tokens. Optical metadata variables are provided as per-tile scalar values, while static-map-derived variables are averaged over the tile footprint to obtain one scalar per field. Each value is assigned to a field-specific interval and represented by a string that identifies both the field and the interval. For example, using longitude bins of width 0.25$^\circ$, a center longitude of 87.49$^\circ$ is assigned to the interval [87.25$^\circ$,87.50$^\circ$) and represented as \texttt{C\_LON=87.25-->87.50}. A text tokenizer then converts the resulting string for each variable into a token ID. Similar to the image case, modality-specific learnable embedding layers map the token sequences to the shared encoder embedding dimension.
 
\end{itemize}

\begin{table*}[ht]
\centering
\caption{Context variables derived from static maps and optical metadata. For each tile, the corresponding map is averaged over the tile footprint and quantized into a single token.}
\label{tab:context-features}
\small
\begin{tabular}{p{2.2cm} p{7.0cm} p{3.8cm}} \toprule
Category & Variables & Description \\ \midrule
Optical \newline metadata &
  \texttt{EM\_ANG}\newline
  \texttt{INC\_ANG}\newline
  \texttt{PHASE\_ANG}\newline
  \texttt{SS\_GROUND\_AZIMUTH}\newline
  \texttt{SS\_LAT}\newline
  \texttt{SS\_LON}\newline
  \texttt{C\_LON}\newline
  \texttt{C\_LAT} &
  - Emission, incidence, and phase angles describing viewing geometry\newline- sub-solar azimuth, latitude, and longitude anchoring the solar reference frame\newline- tile center coordinates \\ \midrule
Photometric \newline properties &
  \texttt{WAC\_NORM\_REF\_[321,360,415,566,604,643,689]}\newline
  \texttt{ALBEDO} &
  Photometrically normalized WAC reflectance products and broadband albedo. \\ \midrule
Illumination and \newline thermophysical &
  \texttt{AVG\_ILLUM}\newline
  \texttt{TBOL\_closest}\newline
  \texttt{TBOL\_POLES\_closest}\newline
  \texttt{HPAR}\newline
  \texttt{TREG} \newline 
  \texttt{DICE} &
  Quantities describing illumination conditions and thermal environment. \\ \midrule
Composition and \newline maturity proxies &
  \texttt{SP\_MINERALOGY\_feo}\newline
  \texttt{SP\_MINERALOGY\_high\_calcium\_pyroxene}\newline
  \texttt{SP\_MINERALOGY\_low\_calcium\_pyroxene}\newline
  \texttt{SP\_MINERALOGY\_nanophase\_iron}\newline
  \texttt{SP\_MINERALOGY\_olivine}\newline
  \texttt{SP\_MINERALOGY\_omat}\newline
  \texttt{SP\_MINERALOGY\_plagioclase}\newline
  \texttt{SW\_FE\_mpfe}\newline
  \texttt{SW\_FE\_npfe}\newline
  \texttt{SW\_FE\_smfe}\newline
  \texttt{TIO2} &
  Composition and maturity proxies from Kaguya-derived spectral mineralogy, space-weathering iron partitioning, and titanium.  \\ \midrule
Volatile-related &
  \texttt{HYDROGEN} &
  Hydrogen abundance proxy. \\ \midrule
Physical and \newline geophysical &
  \texttt{GRAVITY}\newline
  \texttt{ROCK\_ABUNDANCE}\newline
  \texttt{ROUGHNESS} &
  Physical surface characteristics and geophysical products. \\ \bottomrule
\end{tabular}
\end{table*}

\subsection{Multimodal pretraining framework}
\label{sec:pretraining}

Pretraining follows the TerraMind masked-token objective \citep{jakubik2025terramind} (Figure~\ref{fig:fm-overview}, center-right). The encoder can jointly process tokens and patches from the different input modalities. A learnable modality embedding is added to each token or patch, together with a sine-cosine positional embedding: 2D for dense modalities and 1D for sequence modalities. To predict the selected target tokens, the decoder receives the encoder output together with mask tokens for dense modalities or left-shifted tokens for sequence modalities.

At each training step, we draw per-modality token-budget fractions for the input and target sets from separate Dirichlet distributions over the modalities. 

\begin{equation}
f(\mathbf{p}\mid\boldsymbol{\alpha})
= \frac{1}{B(\boldsymbol{\alpha})}
\prod_{m=1}^{M}p_m^{\alpha_m-1},
\qquad
p_m \geq 0,
\qquad
\sum_{m=1}^{M}p_m=1,
\label{eq:dirichlet-allocation}
\end{equation}

where $M$ is the number of eligible modalities in the sample, $B(\boldsymbol{\alpha})$ is the multivariate beta function, $\alpha = (\alpha_1, \dots, \alpha_M)$ is the concentration vector, and $\pi_m$ determine the fractions of the budgets assigned to modality $m$. Eligibility depends on the sample: NAC- and WAC-centered samples contain different modality sets, and only modalities present in that sample can be selected as corresponding inputs or targets.

These fractions are converted into per-modality token counts that can round to zero (absence of modality) if too small. With the per-modality budgets defined, we then sample the tokens within modalities: for dense modalities, patches or tokens are sampled uniformly, and for sequence modalities we use span-masking. This sampling mechanism of allocating input and target tokens exposes the model to a broad range of within- and cross-modal prediction tasks. 

Although the inputs can include tokens and pixel-level patches, the prediction targets are always discrete tokens. Predicting the discrete tokens in a finite vocabulary can be seen as a classification problem over that vocabulary. The decoder outputs a categorical distribution at each target position, and the loss is the standard cross-entropy against the ground-truth codes.

\subsection{Model design considerations}
\label{sec:model-design}

\paragraph{Acquisition geometry as explicit context.}
Standard vision transformers encode spatial arrangement of image patches through positional embeddings but do not necessarily represent acquisition conditions explicitly. This is particularly relevant for lunar optical data, where image characteristics vary with illumination and viewing geometry. Incidence, emission, and phase angles characterize the photometric geometry, while latitude and local solar geometry provide context for illumination, shadowing, and surface temperature. Since this information is available for each tile, we provide it to \lfm as inputs. Explicit conditioning allows the model to account for known acquisition and environmental context when learning representations of the lunar surface.

\paragraph{Mixed low- and high-resolution inputs.}
Each tile is anchored to one of two native optical grids (LROC NAC at $\sim$1~m per pixel or LROC WAC at $\sim$100~m per pixel) with every non-optical modality (i.e. DTM and derived modalities) resampled to its tile's anchor grid. Rather than forcing the corpus onto a single common grid, we let the data-parallel setup mix the two: each worker independently samples either a NAC- or a WAC-anchored batch at each step, so a typical global step carries gradients from both resolution regimes and combines them through the standard all-reduce. Per-modality generated token losses are averaged into one scalar per step that drives a single backward pass, with no separate optimization stream per anchor type. This joint training strategy is intended to expose the model parameters to both resolution regimes while retaining their complementary spatial characteristics.

\paragraph{FlexiViT patch-size adaptation.}
Downstream lunar tasks vary widely in the spatial scale of the phenomena of interest (kilometer-scale craters, meter-scale hazards, sub-tile polar terrain classes). To let a single pretrained checkpoint serve tasks at different working scales, the model adopts the FlexiViT \citep{beyer2023flexivit} patch-embedding scheme: the patch-embedding weights learned at pretraining time can be resized on the fly to a different patch grid at fine-tuning time, without re-training the backbone.

\subsection{Pretraining Setup}
\label{sec:fm-pretraining}

The pretraining procedure comprises three stages: (1) training the modality-specific tokenizers, (2) using the trained tokenizers to convert the pretraining data into discrete tokens, and (3) training \lfm on the resulting tokenized data. Tokenizing the data in advance enables efficient multimodal pretraining by avoiding repeated tokenizer inference. From stage (2) onward, we use 256$\times$256 images (i.e., the model input size) obtained by cropping the corners of the original 512$\times$512 SomBench samples. Corner cropping ensures alignment between the pre-computed tokenized data and the raw modalities, which would not be guaranteed under random cropping without re-running tokenization for each sample at each pretraining step. In the following paragraphs, we describe the hyperparameter configurations used for tokenizer training and \lfm pretraining.

\paragraph{Tokenizers}

We train nine modality-specific tokenizers considering the same encoder-decoder architecture: a ViT-B encoder with a patch size of 16$\times$16 and a U-Net-patched decoder. Latent representations are quantized using FSQ~\cite{fsq} with a single codebook and per-dimension levels of (8, 8, 8, 6, 5), yielding a five-dimensional quantized latent representation.  The generative decoder is trained using a denoising diffusion probabilistic model (DDPM) with 1{,}000 diffusion timesteps and a linear beta schedule.

Optimization is performed using AdamW with $\beta_1 = \text{0.9}$, $\beta_2 = \text{0.99}$, and a weight decay of 0.05, together with a cosine learning-rate schedule that decays to zero and fp16 mixed precision. The training objective combines pixel-wise mean squared error with a perceptual loss computed from intermediate activations of an ImageNet pretrained ViT-Tiny model. The number of GPUs, training epochs, warmup duration, and learning rate vary by modality and is detailed in the Appendix.

\paragraph{Training configuration.}
The encoder is ViT-B with 768 hidden dimensions, 12 encoder layers, and 12 attention heads; the decoder is a Transformer with 12 layers and shared width. Training is carried out on 16 H100 GPUs for 150k steps at a peak learning rate of 1e-4 with a cosine schedule and 15k warm-up steps. Global batch size is 1,536 tiles. Mixed-precision training uses bf16, with the final pretraining run that took 1.1k GPU-hours. The pretraining configuration is released as a yaml file alongside the model weights, and the pretraining data is exactly as released in SomBench \cite{sombench2026collection}.

\subsection{Finetuning Setup}
\label{sec:fm-finetuning}

All finetuning experiments are implemented using TerraTorch \cite{Gomes2025TerraTorch}, an open-source PyTorch-based framework for geospatial foundation-model adaptation. TerraTorch provides a unified interface for assembling encoder--decoder pipelines from interchangeable backbones and task-specific decoder heads (UNet for dense prediction, Faster/Mask R-CNN for detection and instance segmentation, and a ViTDet-style regression head for dense regression), configured via a single YAML file without custom training code. On top of TerraTorch we build a custom module that packages our benchmark datasets and backbone wrappers, adds support for registering new input modalities into the pretrained backbone at fine-tuning time, and implements FlexiViT patch-embedding interpolation (so the pretrained checkpoint can be re-used at working patch scales other than that used in the pretraining). Within each downstream task the data loaders, splits, augmentations, loss, and evaluation metric are held fixed across all backbones so that only the encoder weights and initialization differ. Each configuration is repeated for five random seeds (see per task details below), and we report mean and standard deviation across them. Full hyperparameter configurations are released alongside the finetuning code on \href{https://github.com/NASA-IMPACT/NASA-IBM-Lunar-Foundation-Model}{GitHub}.

We compare our model against a set of widely-used baselines spanning the two dominant architectural families (convolutional and transformer based   ) for geospatial fine-tuning. For the convolutional architectures we include ResNet-50 \citep{he2016deep}, whose skip-connection design remains a well-understood supervised baseline, and the modernized ConvNeXt-Base \citep{liu2022convnet} and its ConvNeXt-V2 successor \citep{woo2023convnextv2}, which incorporate Vision-Transformer design principles while retaining spatial locality. For transformer architectures we include ViT-Base pretrained under the MAE self-supervised objective \citep{dosovitskiy2020image,he2022masked}, together with the hybrid local-global attention variants Swin V2-Base \citep{liu2022swinv2} and DaViT-Base \citep{ding2022davit}. All backbones are initialized from public ImageNet-1k or ImageNet-22k pretrained weights loaded through standard registries. To isolate the contribution of pretraining from that of architecture, we additionally run a from-scratch ResNet-50 control and a randomly initialized copy of our model. For IMP segmentation, which uses a segmentation specific decoder, we also include two task specific architectures assembled through the standard \texttt{segmentation\_models\_pytorch} factory: DeepLabV3+ with a ResNet-50 backbone \citep{chen2018encoder} and SegFormer with a MiT-B2 encoder \citep{xie2021segformer}.

The optimizer recipe is deliberately not shared across backbone families. Our pretrained model uses AdamW with layer-wise learning-rate decay (LLRD) and a lower encoder learning rate than head learning rate. This aims at avoiding that the representations learned during pretraining are pushed away in the early epochs of fine-tuning. All ImageNet-pretrained baselines and the from-scratch controls (including the random-init copy of our model) use AdamW with a flat learning rate and weight-decay values commonly used in the natural-image finetuning literature for their respective architectures. For impact processes and volcanic history tasks we additionally evaluate a parameter-efficient LoRA variant of our model, applying rank-16 adapters ($\alpha=$32) to the attention and MLP linear layers of the pretrained encoder while keeping the remaining encoder weights frozen; the decoder head is trained normally. The task-specific values used per benchmark are given in the paragraphs below.

\paragraph{Impact Processes finetuning configuration.}
All backbones are fine-tuned with a Faster R-CNN head. For our ViT-based encoder we assemble a Multilayer Simple Feature Pyramid neck that reshapes encoder tokens from four selected layers (indices 2, 5, 8, 11 of 12, zero-indexed) into a spatial grid and interpolates them into a four-level FPN before the RCNN heads. Optimization uses AdamW with a cosine schedule after 500 warm-up steps and a floor of $\eta_{\min} = \text{1e-6}$. Our pretrained backbone uses LLRD with backbone and head learning rates both set to 5e-5, layer decay 0.65, encoder weight decay 0.05--0.1 and head weight decay 1e-3, together with stochastic depth (drop\_path = 0.1), following the recipe standard for MAE-pretrained ViTs. The random-init copy of our model uses a flat learning rate (1e-4 on both backbone and head, layer decay 1.0, no stochastic depth), since there are no pretrained features to preserve. All ImageNet baselines use the same flat 1e-4 learning rate on both backbone and head. Anchor sizes are set per-dataset to match crater size distributions (NAC: sizes ranging from 4 to 96 px across four pyramid levels, aspect ratios [0.7, 1.0, 1.4] reflecting near-circular geometry). RPN and ROI sampling budgets are increased (RPN 512 samples/image, ROI 1024 samples/image, 600 detections/image) to accommodate the density of craters per tile. Training data is augmented with D4 group augmentation (rotations and flips) applied jointly to images and bounding boxes; images are 256$\times$256 (NAC, 1~m/px) or 512$\times$512 (WAC, 100~m/px). For WAC we subsample to 50\% of the dataset to assess label efficiency. All models are trained with batch size 4 for up to 100 epochs with early stopping on validation mAP (patience 15), and best checkpoints are selected by validation mAP.

\paragraph{Volcanic History finetuning configuration.}
The IMP segmentation benchmark comprises 100 / 20 / 10 image--mask pairs (train / validation / test) of 256$\times$256 single-channel NAC tiles, and is framed as a binary semantic-segmentation task (background vs.\ Irregular Mare Patch). All backbones use a UNet decoder trained with Dice loss and an auxiliary segmentation head (weight 1.0), with the exception of DeepLabV3+/ResNet-50 and SegFormer/MiT-B2, which use their native segmentation heads and a single unified learning rate. Inputs are standardized with per-dataset statistics computed over the training split after masking no-data pixels, and augmented with D4 group transforms and random brightness/contrast jitter. Every configuration is trained with batch size 32 for a fixed budget of 600 epochs using AdamW with a cosine schedule and 500 warm-up steps; our pretrained backbone uses LLRD 0.85 with backbone learning rate 1e-4 and head learning rate 3e-4, while the ImageNet baselines and from-scratch controls use a flat learning rate of 1e-4. Experiments are repeated for five random seeds.

\paragraph{Polar Volatiles finetuning configuration.}
All ice-prospectivity models share a ViTDet-style Simple Feature Pyramid (SFP) $\rightarrow$ torchvision FPN $\rightarrow$ sum-fuse decoder $\rightarrow$ 4-stage GroupNorm+GELU deep head $\rightarrow$ 1$\times$1 regression conv. For our ViT-B backbone (patch size 8, 32$\times$32 token grid, 12 blocks) we tap four encoder blocks (2, 5, 8, 11) and feed each into a distinct SFP scale head (upsample 4$\times$ / 2$\times$ / identity / MaxPool, yielding strides 4/8/16/32). Each modality has its own pretrained \texttt{PatchedInputAdapter} and tokens are concatenated along the sequence axis, preserving per-modality pretraining. The ImageNet baselines, which cannot do late fusion natively, share the same FPN, decoder and head but use early channel-concatenation of the nine input bands at the stem.  This is done given their lack of ability to process multimodal data natively. The FPN unifies all pyramid levels to 256 channels; the decoder bilinearly upsamples every level to the finest scale, sums them, and further upsamples to the full 256$\times$256 input resolution before a 4-conv GroupNorm head. Optimization uses AdamW ($\beta=(\text{0.9, 0.98})$, $\varepsilon=\text{1e-6}$) with LLRD 0.85 on our pretrained backbone and a flat learning rate on every baseline. Batch size is 4, warm-up over 130 steps ($\approx$10 epochs) is followed by cosine decay to 1e-6; weight decay is 0.05 and gradient norm is clipped at 0.5. The training loss is MSE. Inputs are per-channel z-score standardized and augmented with the D4 group. Every model trains at both backbone and head learning rate 3e-4 for 100 epochs.

For this task, we run two experiments: a full eight-modality comparison of our model (pretrained and randomly initialized) against the ImageNet baselines, repeated across five random seeds; and an ``add-one'' modality-count ablation for our model and ConvNeXt-Base that starts from the two pretrained modalities \{aspect, slope\} and progressively adds one modality at a time in domain-scientist priority order (DICE $\rightarrow$ TMAX $\rightarrow$ LPSR $\rightarrow$ LPSR\_DEN $\rightarrow$ LPSR\_DIS $\rightarrow$ CUR), yielding configurations $m=\text{2}\ldots\text{8}$. This is repeated across three random seeds. Every ablation uses the same architecture, learning-rate schedule and training length as its baseline row, keeping the comparison controlled to the axis being varied.

\section{Results}
\label{sec:experiments}

\subsection{Multimodal correlations exploration}
\label{sec:generations}

Before turning to downstream fine-tuning, we ask a simpler question of the pretrained model: has it learned the cross-modal correspondences that its objective is designed to capture? A masked-token generative model can be probed for this qualitatively by asking it to generate a subset of modalities conditioned on the remaining ones. We stress upfront that the goal of the model is not to serve as a lunar generative product, and that generated fields should not be interpreted as scientific-grade outputs; the examples in this section are qualitative sanity checks on the pretraining objective, not calibrated predictions. A systematic evaluation of generation quality is out of scope for this paper; the downstream evaluations in Section \ref{sec:finetuning_results} test the representation on tasks that have well-defined ground truth and standardized metrics. Here we report several examples of chained generation. In this setting, when multiple outputs are required, each output modality is generated individually and sequentially, with the output of each step used as input to the next. This approach promotes greater coherence across modalities, although errors introduced in earlier steps may propagate to subsequent generations. Additional examples are provided in the Appendix.

\begin{figure}[ht]
    \centering
    \includegraphics[width=0.49\linewidth, trim=0cm 0cm 0cm 0.9cm,
    clip]{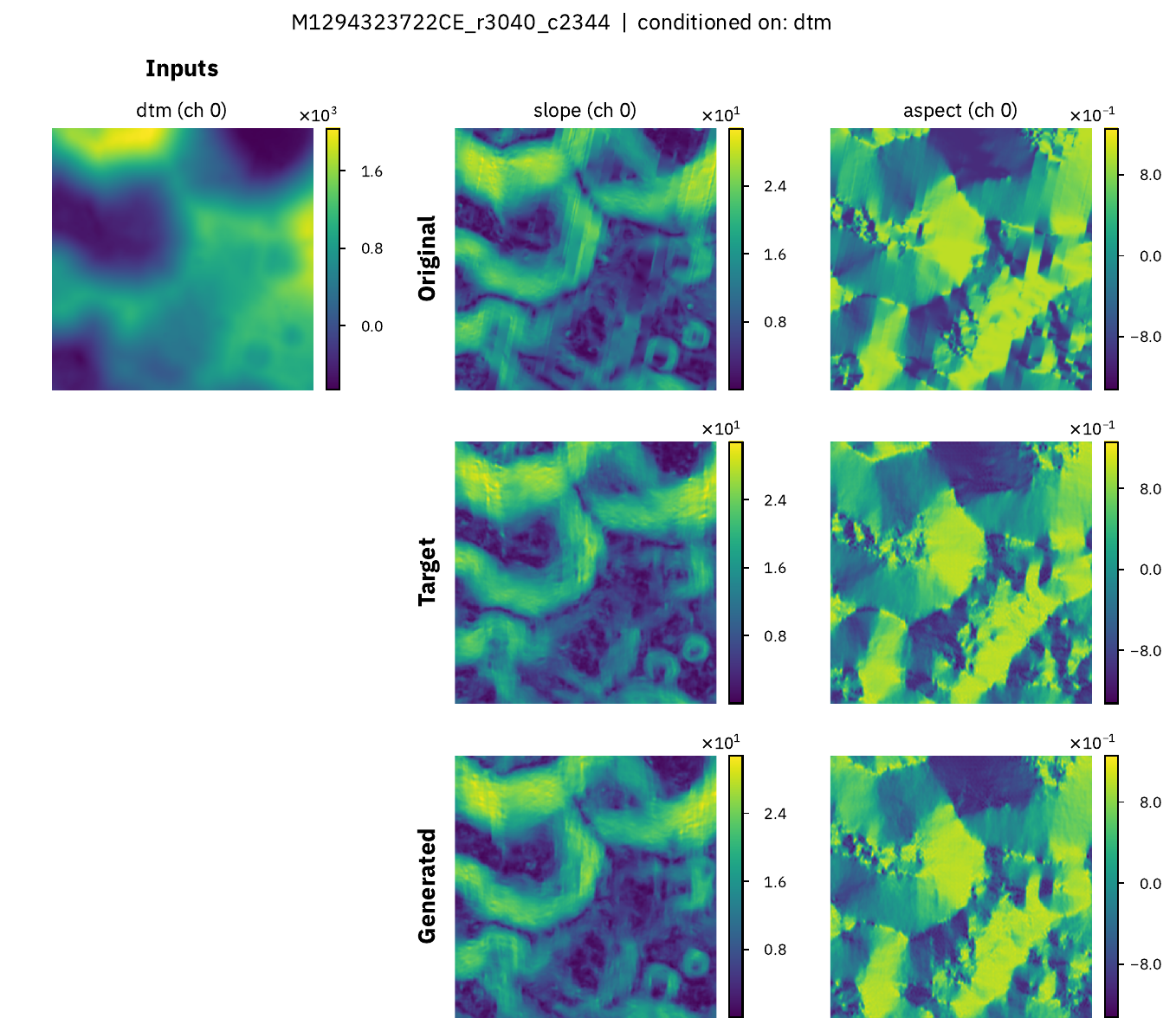}
    \hfill
    \includegraphics[width=0.49\linewidth, trim=0cm 0cm 0cm 0.9cm,
    clip]{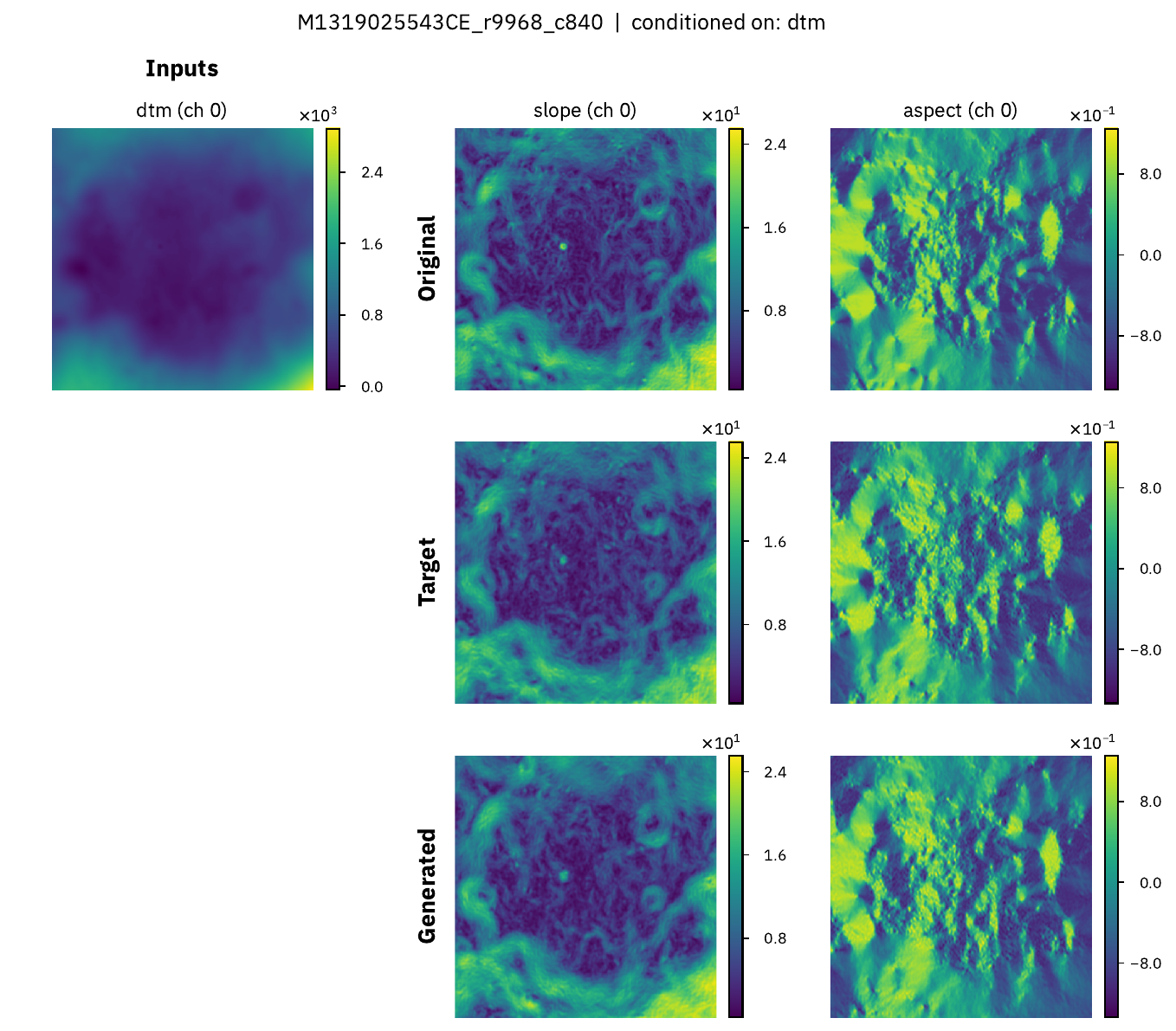}
    \caption{Example of generation from  DTM to slope and aspect for the WAC family.}
    \label{fig:dtm2slope_aspect_examples}
\end{figure}

\begin{figure}[ht]
    \centering
    \includegraphics[width=0.49\linewidth, trim=0cm 0cm 0cm 0.9cm,
    clip]{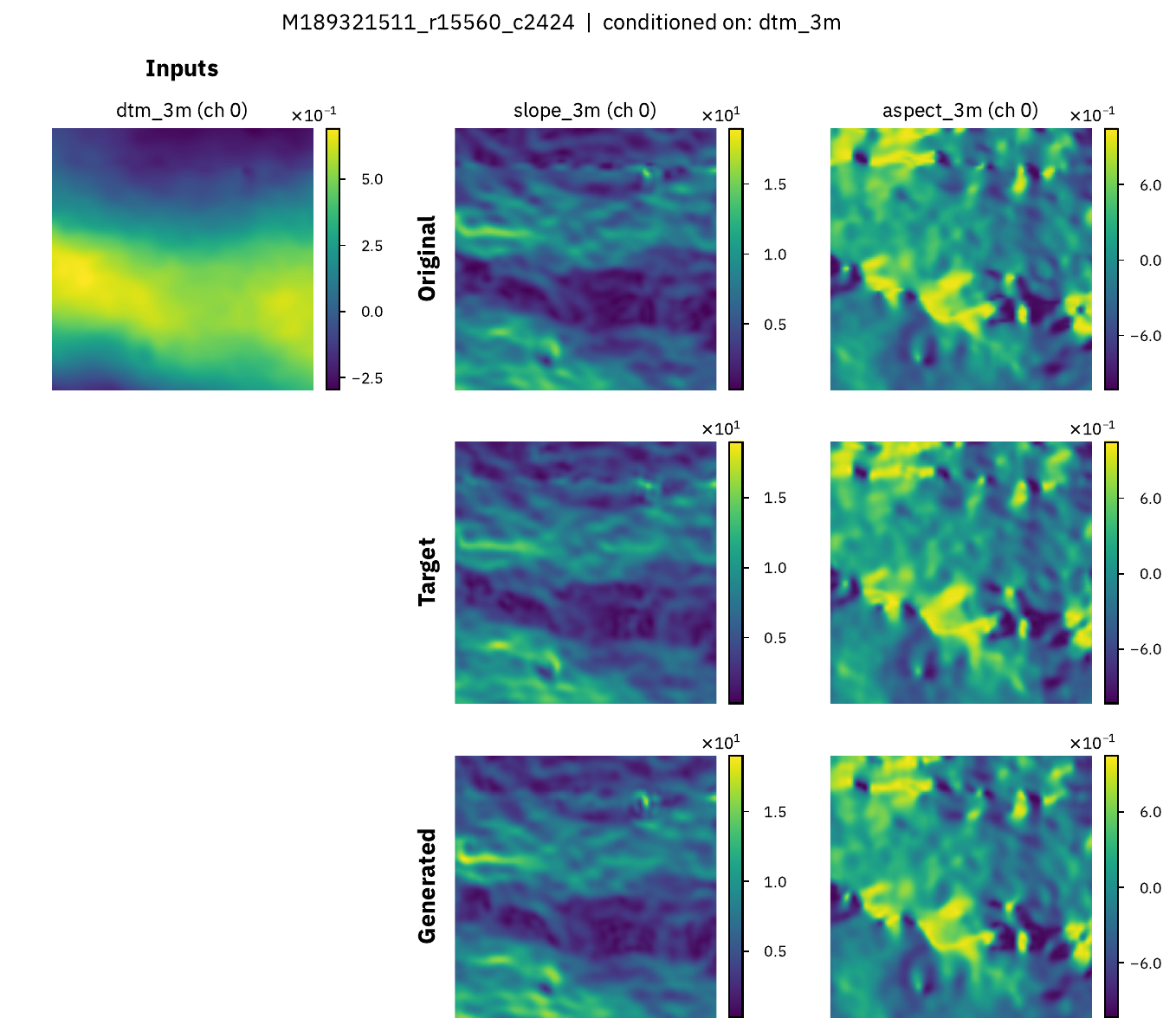}
    \hfill
    \includegraphics[width=0.49\linewidth, trim=0cm 0cm 0cm 0.9cm,
    clip]{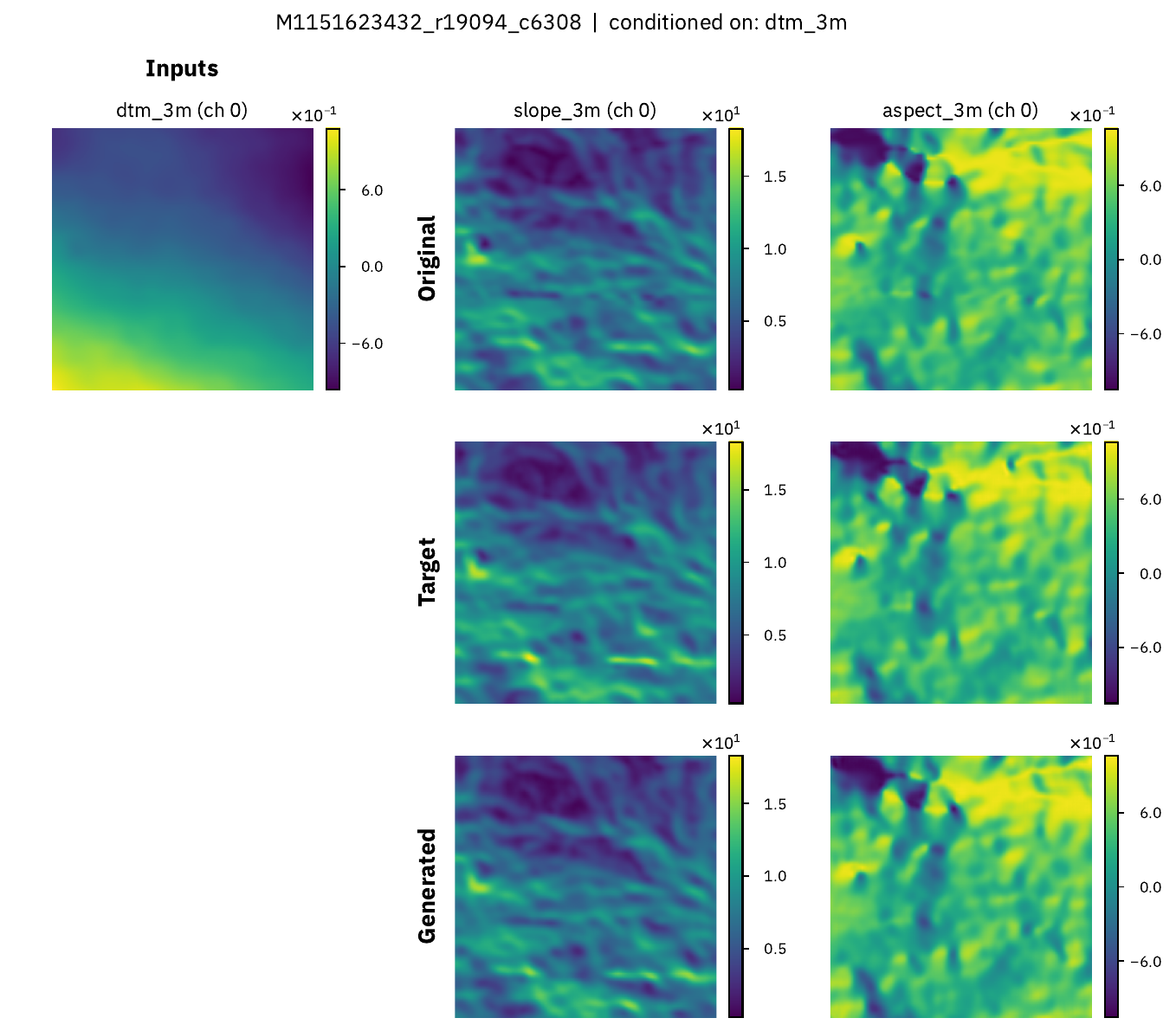}
    \caption{Example of generation from  DTM to slope and aspect for the NAC family.}
    \label{fig:dtm3m2slope3m_aspect3m_examples}
\end{figure}

\paragraph{DTM to slope and aspect.}
The simplest correlation to check is between elevation and its analytical derivatives. Figure \ref{fig:dtm2slope_aspect_examples} and Figure \ref{fig:dtm3m2slope3m_aspect3m_examples} show the model generating WAC-scale and NAC-scale slope and aspect maps conditioned on the corresponding DTM. Slope and aspect are near-deterministic functions of elevation, so we expect the model to have learned this mapping. Qualitatively, this seems the case: generated fields track the target closely at both scales, and the derivative structure (crater rims, ridge orientations, shadowed slopes) is preserved.

\begin{figure}[ht]
    \centering
    \includegraphics[width=\linewidth, trim=0cm 0cm 0cm 0.7cm,
    clip]{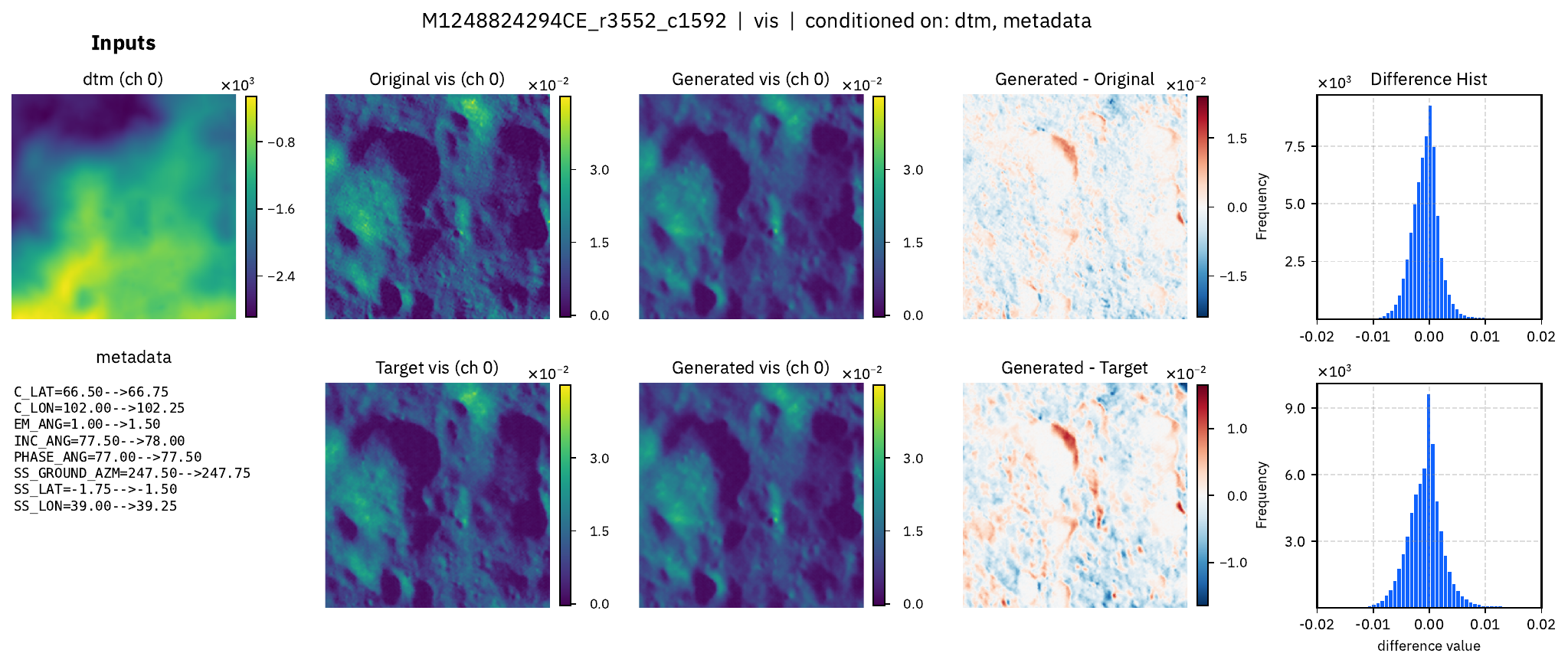}
    \vspace{0.5em}
    \includegraphics[width=\linewidth, trim=0cm 0cm 0cm 0.7cm,
    clip]{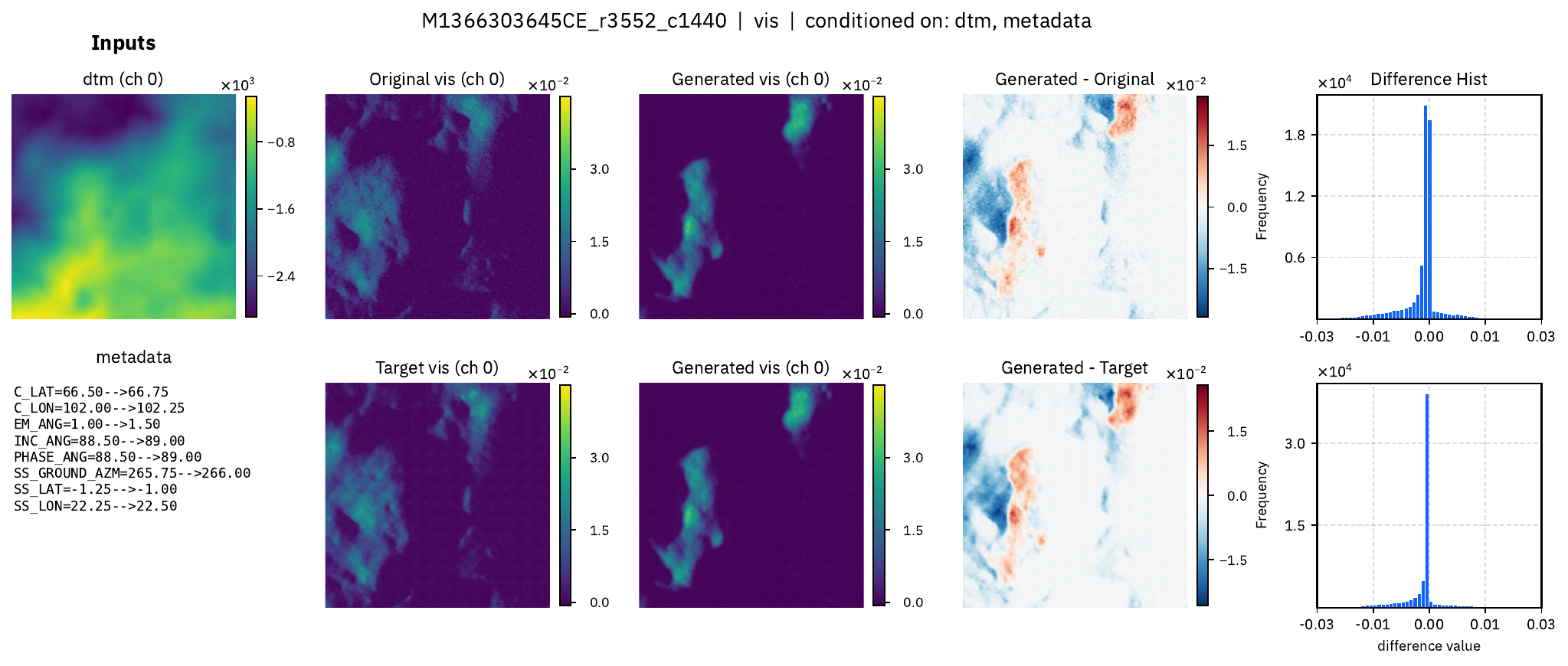}
    \caption{Example of generation from DTM and metadata to WAC for the same location.}
    \label{fig:dtm_metadata2vis_examples}
\end{figure}

\paragraph{DTM and geometry to visible reflectance.}
A stronger test is whether the model can render the same terrain under different acquisition geometries.  Figure \ref{fig:dtm_metadata2vis_examples} shows two WAC visible-reflectance generations at the same geographic location but with different metadata --- one at moderately high incidence, one at a much more grazing incidence. In the first case, the generation is visually close to the observed image, with the difference-histogram concentrating near zero. In the second case, the generated field does not reproduce the target pixel-for-pixel. Nevertheless, the illumination pattern the model produces is qualitatively consistent with the metadata it was given: shadowed regions align with the sun direction implied by the sub-solar coordinates, and the illumination gradient follows the specified geometry. This is the behavior we hoped the sequence-tokenized metadata would elicit, with the conditioning on geometry rather than treating pixel content as the sole signal. However, we caution that this is a single-example qualitative demonstration and not a systematic evaluation of geometric fidelity.

\begin{figure}[ht]
    \centering
    \includegraphics[width=\linewidth, trim=0cm 0cm 0cm 0.9cm,
    clip]{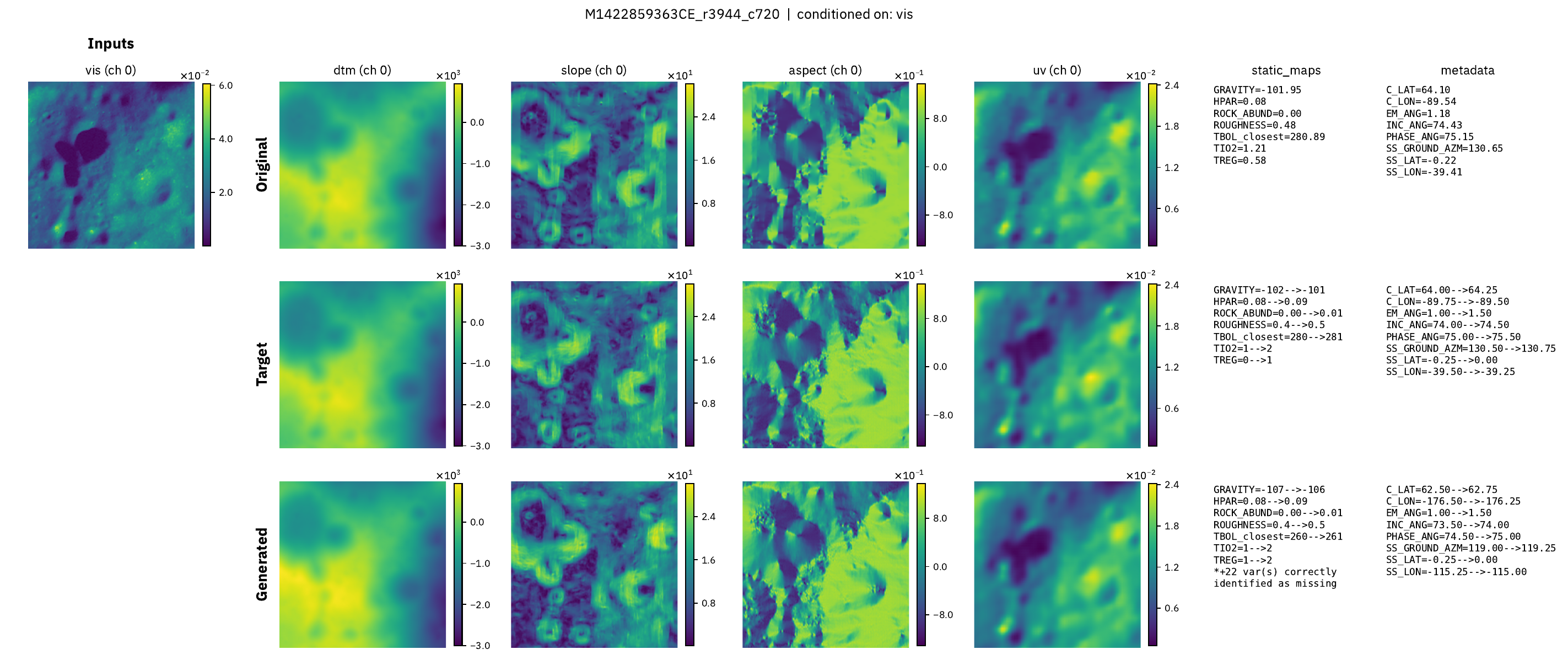}
    \caption{First example of generation from WAC to all other modalities.}
    \label{fig:dtm_metadata2vis_examples1}
\end{figure}

\begin{figure}[ht]
    \centering
    \includegraphics[width=\linewidth, trim=0cm 0cm 0cm 0.9cm,
    clip]{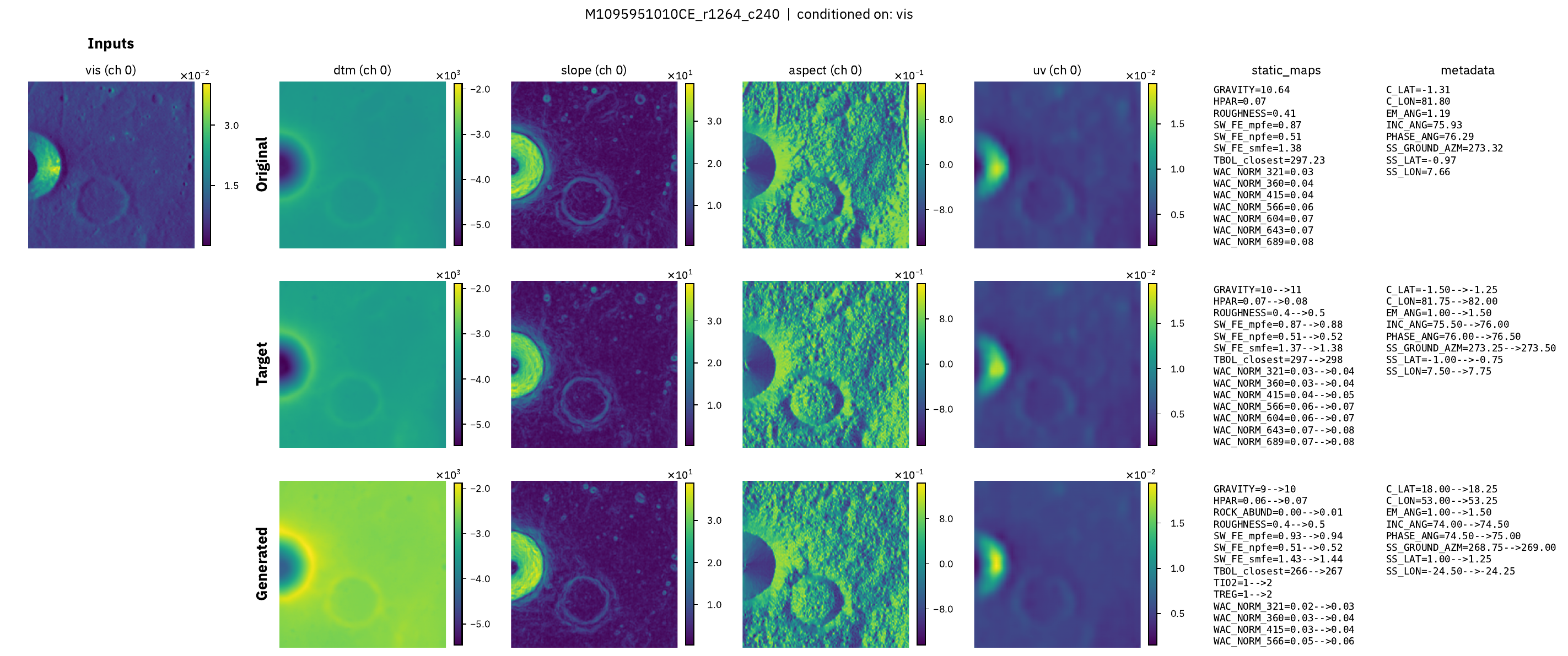}
    \caption{Second example of generation from WAC to all other modalities.}
    \label{fig:dtm_metadata2vis_examples2}
\end{figure}

\paragraph{Chained vis-to-all generation.}
Figure \ref{fig:dtm_metadata2vis_examples1} and Figure \ref{fig:dtm_metadata2vis_examples2} show the harder regime of chained generation across all modalities from a single visible input. In the first example, spatial patterns are recovered across DTM, slope, aspect, and UV, and several static-map scalars fall close to their target values. In the second example, we illustrate a limitation we often observe: the DTM \emph{pattern} is recovered (crater outlines and ridges appear in the right places) but the absolute elevation offset is not, so the generated topography has the correct shape at a shifted value. Similar behavior appears in the static-map bank. Some fields (e.g.\ gravity, roughness, TREG) are close to the target range while others drift more. We observe similar patterns for the metadata output, where absolute latitude and longitude coordinates are frequently off by tens of degrees. These failures are exactly the ones we would expect from a model that has learned local structure from pixel-level correlations but has no direct route to a global geodetic reference frame, and they underline that the model is a representation for downstream tasks rather than a substitute for physical instruments or geodetic solutions.

Taken together, these qualitative examples indicate that pretraining has induced the cross-modal correspondences the objective was designed to capture (i.e., terrain derivatives from elevation, illumination-consistent reflectance from geometry, and coarse spatial pattern across a wide set of co-registered layers), while also making the model's limits legible. The quantitative case for the pretraining recipe is made in Section \ref{sec:finetuning_results}, on downstream tasks with objective evaluation metrics.

\subsection{Finetuning}
\label{sec:finetuning_results}

\paragraph{Impact Processes}

\subparagraph{Robbins Crater Catalog}

Table~\ref{tab:downstream-crater-wac} reports bounding-box mAP on the WAC context-scale crater detection benchmark at two training-data fractions. \lfm (pretrained) leads in every metric column at both fractions: at 50\% training data, full fine-tuning achieves mAP 0.2541$\pm$0.0018 and LoRA achieves mAP 0.2539$\pm$0.0014, both ahead of the strongest ImageNet baseline (SwinV2-B at mAP 0.2313$\pm$0.0027) and of the randomly initialized \lfm (mAP 0.2197$\pm$ 0.0027). At 100\% training data the LoRA variant tops every column (mAP 0.2581$\pm$0.0017, AP@50 0.6183$\pm$0.0034, AP@75 0.2263$\pm$0.0027), while full fine-tuning follows closely at mAP 0.2537$\pm$0.0034. Frozen-backbone fine-tuning underperforms the baselines. The pretrained \lfm variants trained on 50\% of the data already match or exceed SwinV2-B trained on the full set, suggesting labeling efficiency improvement from lunar-domain pretraining. The ranking is stable across data fractions, and the performance gap is substantially larger than the run-to-run variation. This indicates a meaningful advantage of \lfm on this benchmark.

\begin{table}[htpb]
\centering
\caption{Crater detection on LRO WAC (context-scale) as a function of training-data fraction. Test-set bounding-box mean average precision. Values are mean ± standard deviation over five random seeds. \textbf{Bold} indicates the best value per column; \underline{underline} the second best; differences smaller than the reported seed spread should not be read as a ranking.}
\label{tab:downstream-crater-wac}
\begin{tabular}{lccc}
\toprule
Model & mAP & AP@50 & AP@75 \\
\midrule
\multicolumn{4}{c}{\textit{50\% training data}} \\
\midrule
ResNet-50 (ImageNet)                          & 0.1993 $\pm$ 0.0007 & 0.4960 $\pm$ 0.0036 & 0.1572 $\pm$ 0.0038 \\
ResNet-50 (random)                            & 0.1873 $\pm$ 0.0011 & 0.4663 $\pm$ 0.0029 & 0.1472 $\pm$ 0.0041 \\
ViT-B MAE (ImageNet)                          & 0.2167 $\pm$ 0.0022 & 0.5656 $\pm$ 0.0092 & 0.1670 $\pm$ 0.0042 \\
ConvNeXt-B (IN22k)                            & 0.2266 $\pm$ 0.0030 & 0.5684 $\pm$ 0.0030 & 0.1840 $\pm$ 0.0040 \\
ConvNeXtV2-B (IN22k)                          & 0.2303 $\pm$ 0.0034 & 0.5845 $\pm$ 0.0036 & 0.1829 $\pm$ 0.0087 \\
SwinV2-B (ImageNet)                           & 0.2313 $\pm$ 0.0027 & 0.5849 $\pm$ 0.0072 & 0.1862 $\pm$ 0.0041 \\
DaViT-B (ImageNet)                            & 0.2250 $\pm$ 0.0037 & 0.5655 $\pm$ 0.0086 & 0.1781 $\pm$ 0.0099 \\
\midrule
NASA-IBM LFM (ps8, random)                    & 0.2197 $\pm$ 0.0027 & 0.5433 $\pm$ 0.0072 & 0.1781 $\pm$ 0.0038 \\
\midrule
NASA-IBM LFM (ps8)                            & \textbf{0.2541 $\pm$ 0.0018} & \underline{0.6100 $\pm$ 0.0032} & \underline{0.2213 $\pm$ 0.0028} \\
NASA-IBM LFM (ps8, LoRA)                      & \underline{0.2539 $\pm$ 0.0014} & \textbf{0.6103 $\pm$ 0.0012} & \textbf{0.2214 $\pm$ 0.0030} \\
NASA-IBM LFM (ps8, frozen)                    & 0.1617 $\pm$ 0.0026 & 0.3962 $\pm$ 0.0052 & 0.1081 $\pm$ 0.0040 \\
\midrule
\multicolumn{4}{c}{\textit{100\% training data}} \\
\midrule
ResNet-50 (ImageNet)                          & 0.2148 $\pm$ 0.0018 & 0.5288 $\pm$ 0.0055 & 0.1762 $\pm$ 0.0038 \\
ResNet-50 (random)                            & 0.2077 $\pm$ 0.0016 & 0.5131 $\pm$ 0.0017 & 0.1698 $\pm$ 0.0035 \\
ViT-B MAE (ImageNet)                          & 0.2259 $\pm$ 0.0054 & 0.5834 $\pm$ 0.0151 & 0.1785 $\pm$ 0.0078 \\
ConvNeXt-B (IN22k)                            & 0.2308 $\pm$ 0.0067 & 0.5830 $\pm$ 0.0054 & 0.1890 $\pm$ 0.0137 \\
ConvNeXtV2-B (IN22k)                          & 0.2396 $\pm$ 0.0037 & 0.6025 $\pm$ 0.0056 & 0.2014 $\pm$ 0.0080 \\
SwinV2-B (ImageNet)                           & 0.2420 $\pm$ 0.0047 & 0.6020 $\pm$ 0.0061 & 0.2028 $\pm$ 0.0086 \\
DaViT-B (ImageNet)                            & 0.2347 $\pm$ 0.0029 & 0.5981 $\pm$ 0.0051 & 0.1912 $\pm$ 0.0053 \\
\midrule
NASA-IBM LFM (ps8, random)                    & 0.2289 $\pm$ 0.0037 & 0.5595 $\pm$ 0.0080 & 0.1893 $\pm$ 0.0073 \\
\midrule
NASA-IBM LFM (ps8)                            & \underline{0.2537 $\pm$ 0.0034} & \underline{0.6156 $\pm$ 0.0024} & \underline{0.2174 $\pm$ 0.0072} \\
NASA-IBM LFM (ps8, LoRA)                      & \textbf{0.2581 $\pm$ 0.0017} & \textbf{0.6183 $\pm$ 0.0034} & \textbf{0.2263 $\pm$ 0.0027} \\
NASA-IBM LFM (ps8, frozen)                    & 0.1871 $\pm$ 0.0024 & 0.4583 $\pm$ 0.0055 & 0.1317 $\pm$ 0.0038 \\
\bottomrule
\end{tabular}
\end{table}

\subparagraph{NAC Hand Labeled dataset}

Table~\ref{tab:downstream-crater-nac} reports mAP on the meter-scale NAC crater detection benchmark, which generally shows low scores across the board. \lfm finetuned with LoRA achieves 0.1543$\pm$0.0098 mAP, immediately after the best performing model (SwinV2-B) at 0.1552$\pm$0.0086 mAP. DaViT-B performs best for AP@50 and SwinV2-B shows top performance for AP@75, with \lfm following closely. The random-init LFM (mAP 0.1274$\pm$0.0151) is one of the weakest results. The frozen variant (mAP 0.1295$\pm$0.0038) performs similarly to random init on this task, indicating that some encoder adaptation is needed at meter scale for this task. Variability across the different seeds is wide on this task, with performance models at the top that are within each other's standard deviation. This suggests comparable performance among those, rather than a clear winner.  

\begin{table}[htpb]
\centering
\caption{Crater detection on LRO NAC (meter-scale) at 100\% training data. Test-set bounding-box mean average precision. Values are mean ± standard deviation over five random seeds. \textbf{Bold} indicates the best value per column; \underline{underline} the second best; differences smaller than the reported seed spread should not be read as a ranking.}
\label{tab:downstream-crater-nac}
\begin{tabular}{lccc}
\toprule
Model & mAP & AP@50 & AP@75 \\
\midrule
ResNet-50 (ImageNet)                          & 0.1411 $\pm$ 0.0036 & 0.4374 $\pm$ 0.0196 & 0.0895 $\pm$ 0.0055 \\
ResNet-50 (random)                            & 0.1296 $\pm$ 0.0034 & 0.4159 $\pm$ 0.0171 & 0.0787 $\pm$ 0.0057 \\
ViT-B MAE (ImageNet)                          & 0.1236 $\pm$ 0.0119 & 0.3798 $\pm$ 0.0460 & 0.0703 $\pm$ 0.0109 \\
ConvNeXt-B (IN22k)                            & 0.1464 $\pm$ 0.0060 & 0.4400 $\pm$ 0.0249 & 0.0933 $\pm$ 0.0039 \\
ConvNeXtV2-B (IN22k)                          & 0.1345 $\pm$ 0.0085 & 0.3949 $\pm$ 0.0545 & 0.0783 $\pm$ 0.0107 \\
SwinV2-B (ImageNet)                           & \textbf{0.1552 $\pm$ 0.0086} & \underline{0.4586 $\pm$ 0.0308} & \textbf{0.1090 $\pm$ 0.0146} \\
DaViT-B (ImageNet)                            & 0.1501 $\pm$ 0.0096 & \textbf{0.4620 $\pm$ 0.0265} & 0.0983 $\pm$ 0.0080 \\
\midrule
NASA-IBM LFM (ps8, random)                    & 0.1274 $\pm$ 0.0151 & 0.3632 $\pm$ 0.0439 & 0.0818 $\pm$ 0.0139 \\
\midrule
NASA-IBM LFM (ps8)                            & 0.1460 $\pm$ 0.0264 & 0.4565 $\pm$ 0.0732 & 0.0953 $\pm$ 0.0318 \\
NASA-IBM LFM (ps8, LoRA)                      & \underline{0.1543 $\pm$ 0.0098} & 0.4416 $\pm$ 0.0496 & \underline{0.1072 $\pm$ 0.0116} \\
NASA-IBM LFM (ps8, frozen)                    & 0.1295 $\pm$ 0.0038 & 0.4098 $\pm$ 0.0121 & 0.0756 $\pm$ 0.0040 \\
\bottomrule
\end{tabular}
\end{table}

\begin{figure}[t]
    \centering
    \begin{subfigure}{0.45\textwidth}
        \includegraphics[width=\linewidth]{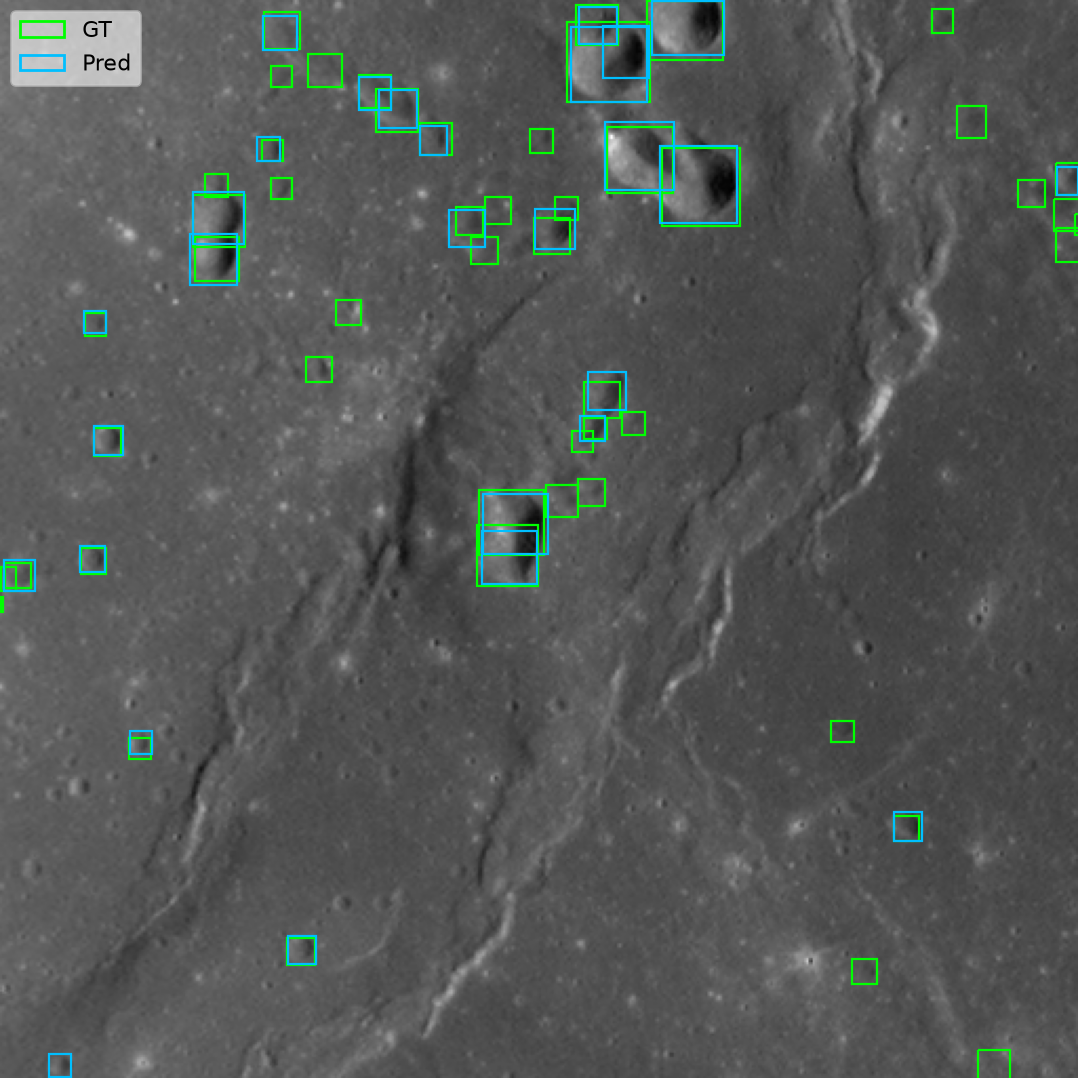}
        \caption{}
    \end{subfigure}
    \hfill
    \begin{subfigure}{0.45\textwidth}
        \includegraphics[width=\linewidth]{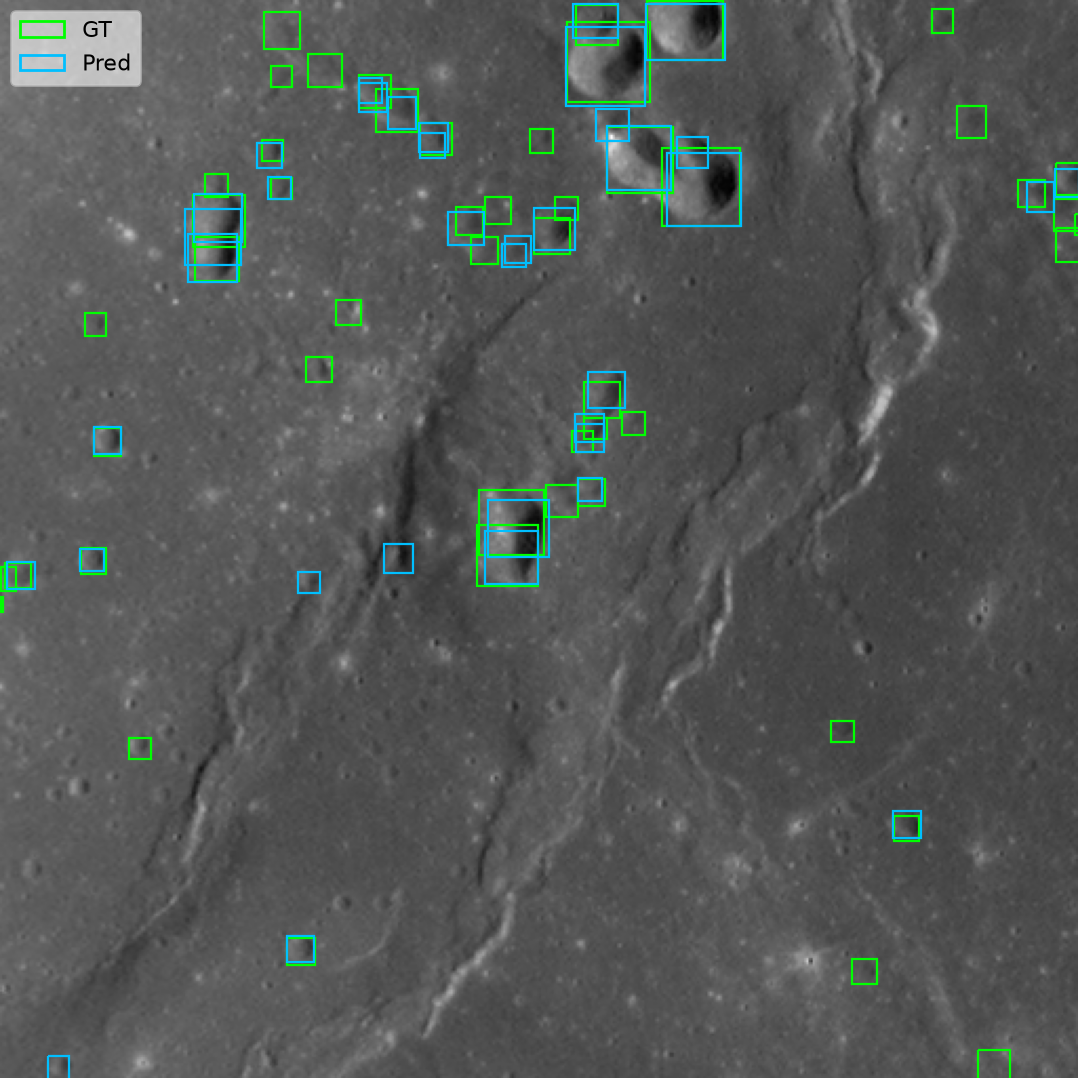}
        \caption{}
    \end{subfigure}

    \vspace{6pt}

    \begin{subfigure}{0.45\textwidth}
        \includegraphics[width=\linewidth]{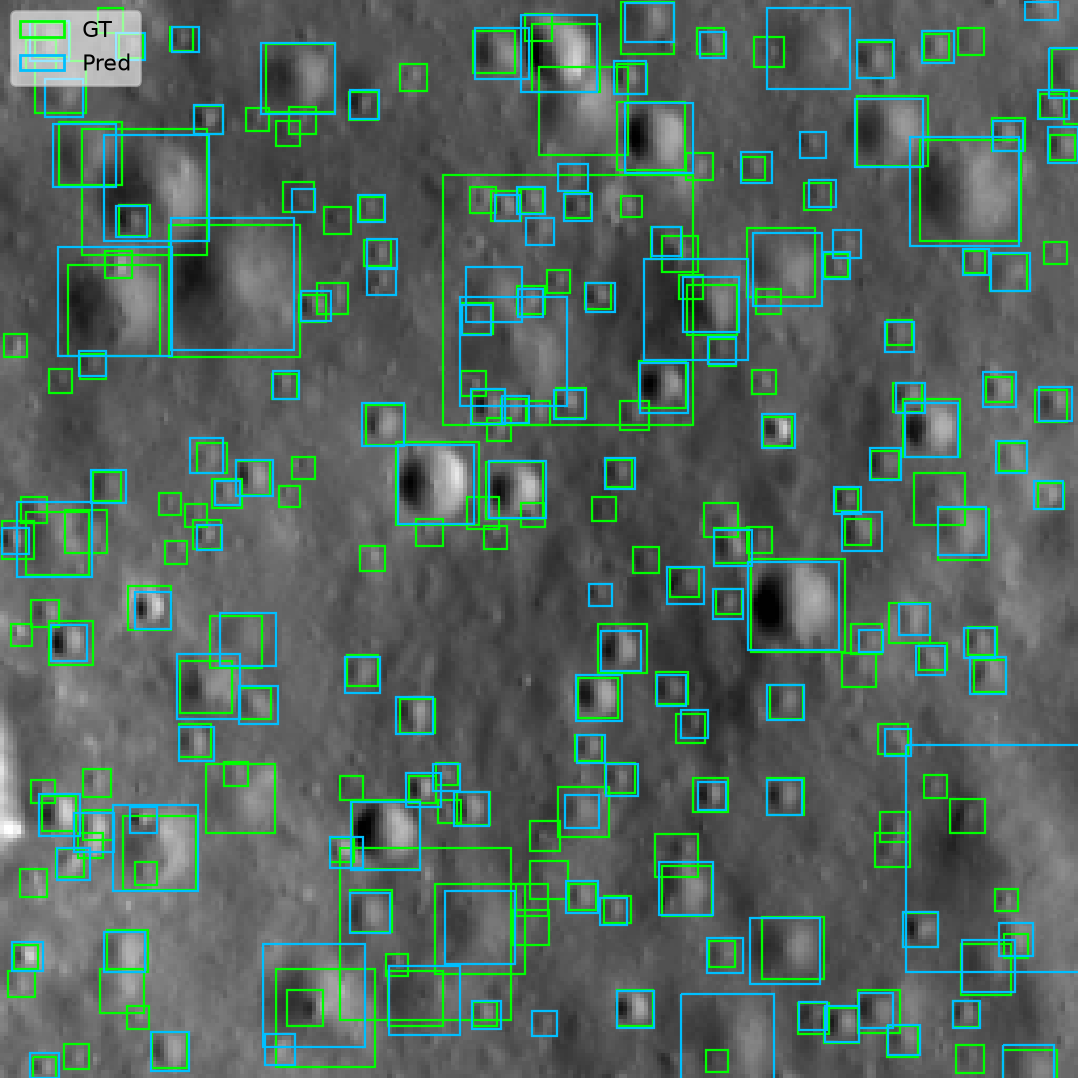}
        \caption{}
    \end{subfigure}
    \hfill
    \begin{subfigure}{0.45\textwidth}
        \includegraphics[width=\linewidth]{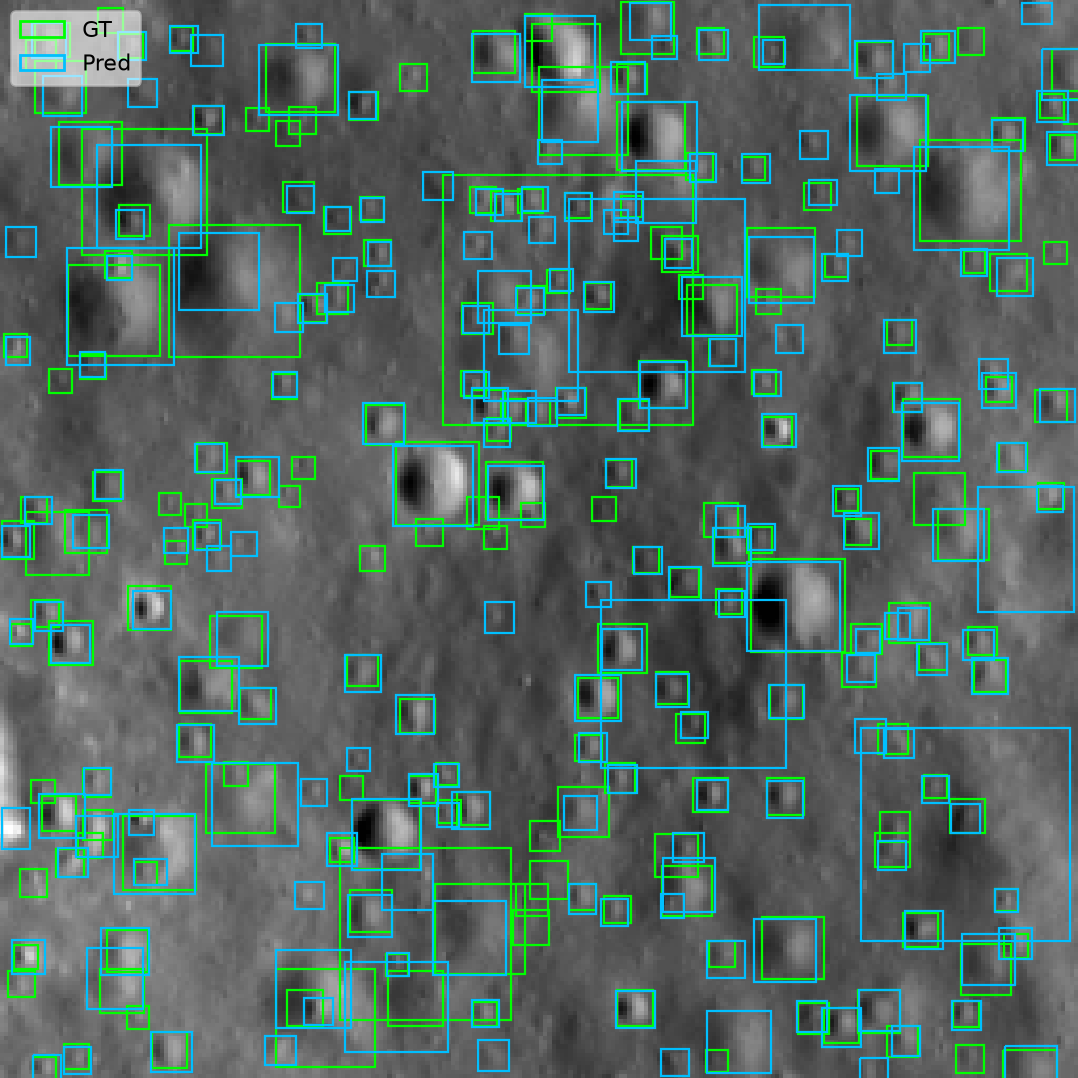}
        \caption{}
    \end{subfigure}

    \caption{Example of predictions for the WAC Robbins Crater Catalog (first row) and the NAC Hand Labeled Dataset (second row). The first column (\textit{a} and \textit{c}) shows predictions over ground truth for \lfm, with the second column (\textit{b} and \textit{d}) showing the same for SwinV2-B. Green: ground truth; blue: predictions at confidence $\geq$ 0.5.}
    \label{fig:craters-examples}
\end{figure}

Figure \ref{fig:craters-examples} presents examples of predictions for a sample for the Robbins crater catalog (top) and NAC Hand Labeled dataset (bottom) for \lfm and SwinV2-B. The predictions for the two models look similar. However the images show some differences across the two datasets. The Robbins catalog presents less dense and more clearly defined labels compared to the NAC Hand Labeled dataset. Furthermore, part of the NAC benchmark was also annotated at 5 m/pixels, and those sites — including the one shown in the bottom row — are visibly blurrier. These two differences might explain the lower scores for all models in the NAC Hand Labeled dataset.

\paragraph{Volcanic history}
\leavevmode\\
\leavevmode\\
Table~\ref{tab:downstream-imp} reports foreground IoU$_1$ and F1$_1$ on the IMP segmentation benchmark. The \lfm variants occupy the top positions, with the frozen-backbone configuration that achieves the best IoU$_1$ (0.5709$\pm$0.0114) and F1 (0.7268$\pm$0.0093), followed by full fine-tuning (IoU$_1$ 0.5693$\pm$0.0114, F1 0.7255$\pm$0.0093). These are slightly ahead of the strongest ImageNet baseline, such as ConvNeXt-V2-B (IoU$_1$ 0.5687$\pm$0.0181 and F1$_1$ 0.7249$\pm$0.0146). \lfm LoRA remains competitive also for this task (IoU$_1$ 0.5593$\pm$0.0137, F1 0.7173$\pm$0.0113), while  the random-init control shows poor performance (IoU$_1$ 0.3142$\pm$0.0746), confirming that pretraining is essential for this low-data segmentation task. While a ranking can be established, the performance of the top models falls within the variability observed across runs, indicating that they achieve comparable results. 

\begin{figure}[ht]
    \centering
    \includegraphics[width=\linewidth, trim=0cm 0cm 0cm 0.9cm,
    clip]{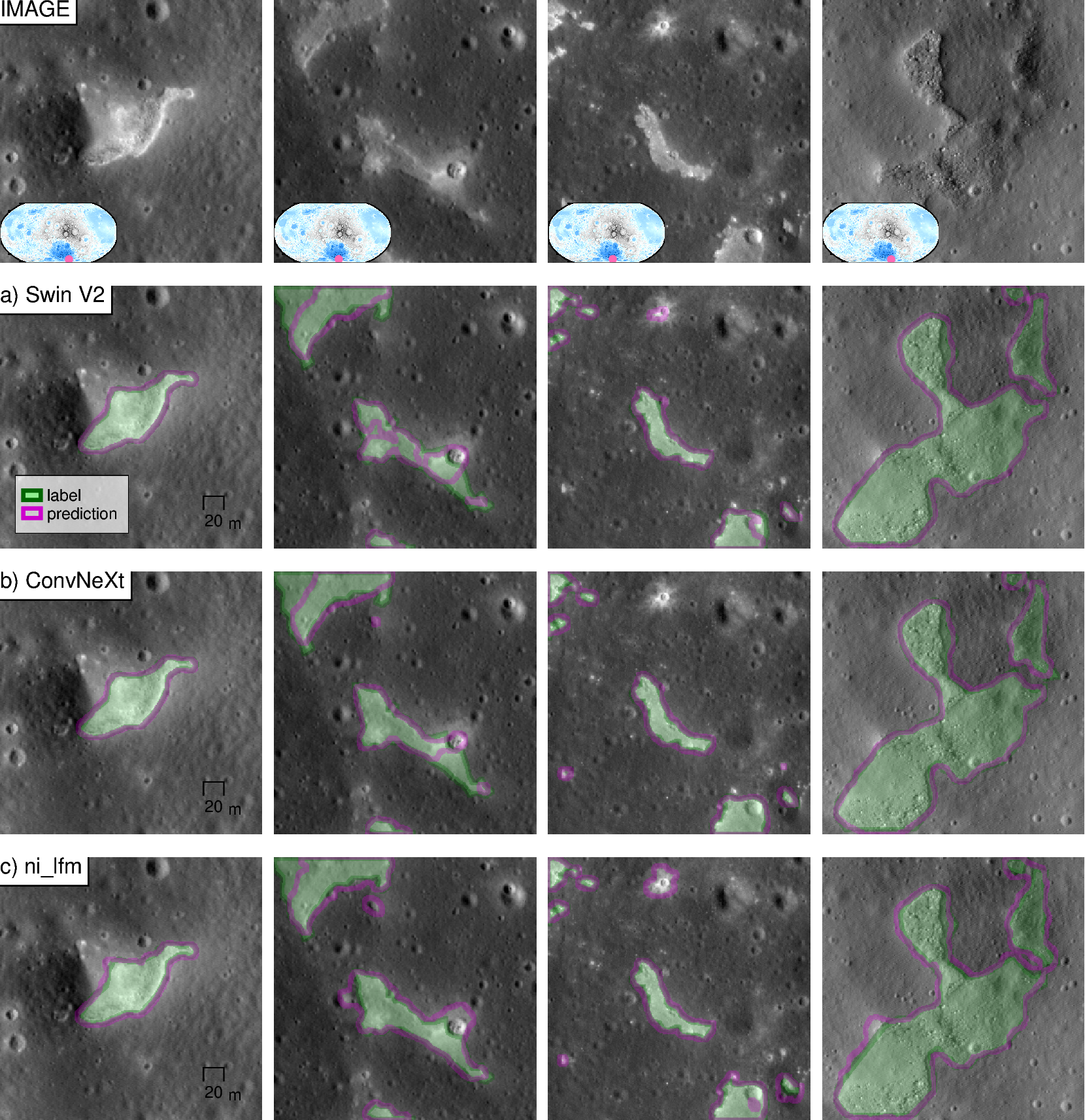}
    \caption{Qualitative IMP segmentation predictions. Our model and the baselines produce broadly comparable outputs; in the second example, the pretrained \lfm identifies IMP extents that none of the baselines detect. Green overlays mark ground-truth IMP extents; colored overlays show model predictions. The colors used in the maps in the top row are for visualization purposes only and do not convey any specific meaning.}
    \label{fig:imp}
\end{figure}

Figure~\ref{fig:imp} shows representative qualitative predictions on four IMP examples. Predictions are broadly similar across models, with \lfm generally identifying slightly larger IMP regions than the baselines, which might lead to more false positives. In the second example, this behavior enables our model to capture IMP regions that are not detected by the baseline models.

\begin{table}[htpb]
\centering
\caption{Irregular Mare Patch (IMP) segmentation. IMP delineation is a binary dense-prediction task, so we report test-set metrics on the positive (IMP) class only: intersection-over-union (IoU$_1$) and F1 score (F1$_1$). Values are mean $\pm$ standard deviation over five random seeds. \textbf{Bold} indicates the best mean per column and \underline{underline} the second best; differences smaller than the reported seed spread should not be read as a ranking.}
\label{tab:downstream-imp-5seeds}
\label{tab:downstream-imp}
\setlength{\tabcolsep}{6pt}
\begin{tabular}{l cc}
\toprule
Model & IoU$_1$ & F1$_1$ \\
\midrule
ResNet-50 (ImageNet)                          & 0.4731 $\pm$ 0.0114 & 0.6423 $\pm$ 0.0105 \\
ResNet-50 (random)                            & 0.3496 $\pm$ 0.0323 & 0.5175 $\pm$ 0.0355 \\
ViT-B MAE (ImageNet)                          & 0.5541 $\pm$ 0.0369 & 0.7125 $\pm$ 0.0298 \\
ConvNeXt-B (IN22k)                            & 0.5374 $\pm$ 0.0197 & 0.6989 $\pm$ 0.0168 \\
ConvNeXtV2-B (IN22k)                          & 0.5687 $\pm$ 0.0181 & 0.7249 $\pm$ 0.0146 \\
SwinV2-B (ImageNet)                           & 0.5555 $\pm$ 0.0159 & 0.7141 $\pm$ 0.0132 \\
DaViT-B (ImageNet)                            & 0.5623 $\pm$ 0.0055 & 0.7198 $\pm$ 0.0045 \\
DeepLabV3+ - ResNet-50                        & 0.5154 $\pm$ 0.0180 & 0.6801 $\pm$ 0.0157 \\
SegFormer-MiT-B2                              & 0.5475 $\pm$ 0.0099 & 0.7076 $\pm$ 0.0083 \\
\midrule
NASA-IBM LFM (ps8, random)                    & 0.3142 $\pm$ 0.0746 & 0.4742 $\pm$ 0.0892 \\
\midrule
NASA-IBM LFM (ps8)                            & \underline{0.5693 $\pm$ 0.0114} & \underline{0.7255 $\pm$ 0.0093} \\
NASA-IBM LFM (ps8, LoRA)                      & 0.5593 $\pm$ 0.0137 & 0.7173 $\pm$ 0.0113 \\
NASA-IBM LFM (ps8, frozen)                    & \textbf{0.5709 $\pm$ 0.0114} & \textbf{0.7268 $\pm$ 0.0093} \\
\bottomrule
\end{tabular}
\end{table}

\paragraph{Polar volatiles}
\leavevmode\\
\leavevmode\\
Table~\ref{tab:downstream-ice-prosp} reports RMSE, MAE, and $R^{2}$ on the full eight-modality ice prospectivity regression benchmark. \lfm (pretrained, full fine-tuning) achieves the lowest RMSE (0.0293$\pm$0.0013, $R^{2} = \text{0.9884}\pm\text{0.0010}$), followed by the LoRA variant (RMSE 0.0330$\pm$0.0005) and the frozen variant (RMSE 0.0357$\pm$0.0007). The best ImageNet-pretrained baseline, SwinV2-B, reaches RMSE 0.0377$\pm$0.0004; ConvNeXt-Base follows at RMSE 0.0442 $\pm$0.0008. The pretrained \lfm variants outperform all ImageNet baselines and the random-init control (RMSE 0.0397$\pm$ 0.0004) across every metric, with a consistent gap well beyond variability across runs.  Table~\ref{tab:downstream-ice-prosp-ablation} reports an ablation in which we start from the two modalities seen during pretraining (aspect and slope) and progressively add polar-stack inputs that neither model ever saw during pretraining. We compare \lfm against ConvNeXt-B, as one of the strongest convolutional baselines. \lfm obtains lower RMSE at every modality count, and the margin is most striking in the sparse-input regime: with only three modalities \lfm reaches RMSE 0.0434$\pm$0.0006, matching ConvNeXt-B using the full eight-modality stack (0.0437$\pm$0.0002). At $m{=}$2 both collapse to RMSE $\approx$0.2, indicating that slope and aspect alone are insufficient for this task regardless of pretraining.

\begin{table}[htpb]
\centering
\caption{Polar ice prospectivity regression on the full eight-modality polar stack (aspect, slope, DICE, TMAX, LPSR, LPSR\_DEN, LPSR\_DIS, CUR). Test-set metrics against the prospectivity map in \cite{coyan2025prospectivity}: RMSE and MAE lower-is-better, $R^{2}$ higher-is-better. Values are mean ± standard deviation over five random seeds. \textbf{Bold} indicates the best value per column; \underline{underline} the second best.}
\label{tab:downstream-ice-prosp}
\begin{tabular}{lccc}
\toprule
Model & RMSE $\downarrow$ & MAE $\downarrow$ & $R^{2}$ $\uparrow$ \\
\midrule
ResNet-50 (ImageNet)                          & 0.0788 $\pm$ 0.0004 & 0.0568 $\pm$ 0.0004 & 0.9162 $\pm$ 0.0009 \\
ResNet-50 (random)                            & 0.0695 $\pm$ 0.0007 & 0.0495 $\pm$ 0.0005 & 0.9347 $\pm$ 0.0013 \\
ViT-B MAE (ImageNet)                          & 0.0917 $\pm$ 0.0009 & 0.0674 $\pm$ 0.0007 & 0.8865 $\pm$ 0.0021 \\
ConvNeXt-B (IN22k)                            & 0.0442 $\pm$ 0.0008 & 0.0301 $\pm$ 0.0005 & 0.9737 $\pm$ 0.0009 \\
ConvNeXtV2-B (IN22k)                          & 0.0439 $\pm$ 0.0008 & 0.0298 $\pm$ 0.0006 & 0.9739 $\pm$ 0.0010 \\
SwinV2-B (ImageNet)                           & 0.0377 $\pm$ 0.0004 & 0.0256 $\pm$ 0.0003 & 0.9808 $\pm$ 0.0004 \\
DaViT-B (ImageNet)                            & 0.0469 $\pm$ 0.0007 & 0.0321 $\pm$ 0.0005 & 0.9704 $\pm$ 0.0009 \\
\midrule
NASA-IBM LFM (ps8, random)                    & 0.0397 $\pm$ 0.0004 & 0.0271 $\pm$ 0.0003 & 0.9787 $\pm$ 0.0005 \\
\midrule
NASA-IBM LFM (ps8)                            & \textbf{0.0293 $\pm$ 0.0013} & \textbf{0.0197 $\pm$ 0.0009} & \textbf{0.9884 $\pm$ 0.0010} \\
NASA-IBM LFM (ps8, LoRA)                      & \underline{0.0330 $\pm$ 0.0005} & \underline{0.0224 $\pm$ 0.0004} & \underline{0.9853 $\pm$ 0.0004} \\
NASA-IBM LFM (ps8, frozen)                    & 0.0357 $\pm$ 0.0007 & 0.0246 $\pm$ 0.0005 & 0.9828 $\pm$ 0.0007 \\
\bottomrule
\end{tabular}
\end{table}

\begin{table}[htpb]
\centering
\small
\caption{Polar ice prospectivity regression, modality-count ablation. Each column corresponds to results from models trained on a fixed set of polar input variables, starting from the two pretrained modalities (aspect, slope) at $m{=}2$ and incrementally adding polar-stack modalities not seen during pretraining, in the order DICE, TMAX, LPSR, LPSR\_DEN, LPSR\_DIS, and finally CUR at $m{=}8$. This isolates the contribution of each additional benchmark-only modality on top of the pretrained backbone. Values are test-set RMSE (lower is better). Values are mean ± standard deviation over three random seeds. \textbf{Bold} indicates the best value per column.}
\label{tab:downstream-ice-prosp-ablation}
\setlength{\tabcolsep}{4pt}
\begin{tabular}{lccccccc}
\toprule
 & \multicolumn{7}{c}{Number of input modalities $m$} \\
\cmidrule(lr){2-8}
Model & 2 & 3 & 4 & 5 & 6 & 7 & 8 \\
 & \scriptsize(as, sl) & \scriptsize+DICE & \scriptsize+TMAX & \scriptsize+LPSR & \scriptsize+DEN & \scriptsize+DIS & \scriptsize+CUR \\
\midrule
ConvNeXt-B (IN22k)        & \makecell{0.2083 \\ {\scriptsize ($\pm$0.0086)}} & \makecell{0.0614 \\ {\scriptsize ($\pm$0.0007)}} & \makecell{0.0522 \\ {\scriptsize ($\pm$0.0007)}} & \makecell{0.0513 \\ {\scriptsize ($\pm$0.0004)}} & \makecell{0.0509 \\ {\scriptsize ($\pm$0.0005)}} & \makecell{0.0517 \\ {\scriptsize ($\pm$0.0007)}} & \makecell{0.0437 \\ {\scriptsize ($\pm$0.0002)}} \\
\addlinespace[0.6em]
NASA-IBM LFM (ps8)        & \makecell{\textbf{0.1981} \\ {\scriptsize ($\pm$0.0063)}} & \makecell{\textbf{0.0434} \\ {\scriptsize ($\pm$0.0006)}} & \makecell{\textbf{0.0412} \\ {\scriptsize ($\pm$0.0009)}} & \makecell{\textbf{0.0393} \\ {\scriptsize ($\pm$0.0002)}} & \makecell{\textbf{0.0403} \\ {\scriptsize ($\pm$0.0010)}} & \makecell{\textbf{0.0398} \\ {\scriptsize ($\pm$0.0004)}} & \makecell{\textbf{0.0288} \\ {\scriptsize ($\pm$0.0011)}} \\
\bottomrule
\end{tabular}
\end{table}

\begin{figure}[ht]
    \centering
    \includegraphics[width=\linewidth, trim=0cm 0cm 0cm 0.9cm,
    clip]{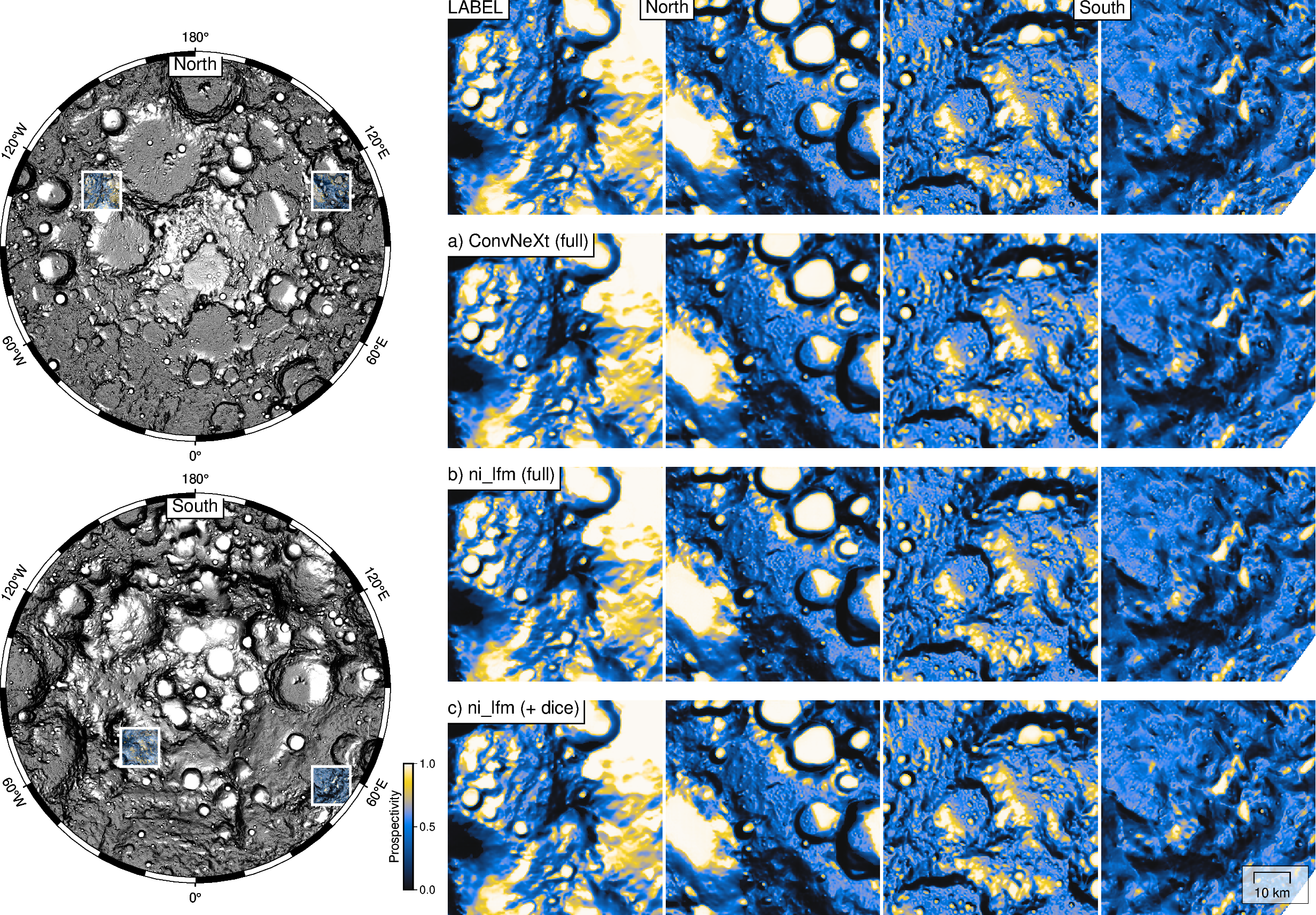}
    \caption{Qualitative ice prospectivity predictions on test patches. \lfm recovers PSR-boundary sharpness and fine spatial detail that the ImageNet-pretrained baseline smooths out. The last row shows our model prediction using only slope, aspect, and DICE, illustrating competitive performance even with a drastically reduced input set.}
    \label{fig:ice_prosp}
\end{figure}

Figure \ref{fig:ice_prosp} shows some predictions for polar volatiles across the south and north pole. The \lfm is able to more accurately preserve details across the different examples compared to ConvNext, even when only a limited number of modalities is used.

\section{Conclusion}
We have presented a multimodal lunar foundation model built by adapting the TerraMind recipe to the lunar setting and pretraining from scratch on SomBench, a co-registered corpus of nearly two million tile bundles spanning 11 modalities across two spatial scales. Beyond the different data used, two lunar-specific extensions distinguish our model from its Earth-observation ancestor: per-tile acquisition geometry is provided as explicit context input, removing a dominant confound from the pixel-level signal; and NAC- and WAC-anchored tiles are trained jointly in a mixed-batch regime, so a single set of weights absorbs both meter-scale geomorphology and hundred-meter-scale regional context. A FlexiViT patch-embedding scheme then lets the same pretrained checkpoint serve downstream tasks at multiple working patch grids without any backbone retraining.

Across four downstream benchmarks the pretrained model matches or exceeds ImageNet-pretrained baselines while also improving over an architecturally identical random-initialization control. The gains are largest where labeled data is scarce: at 50\% training data on the WAC crater benchmark, our pretrained variants already exceed the strongest ImageNet baseline trained on the full set. Furthermore, polar ice prospectivity shows the widest margin of all (RMSE 0.0293$\pm$0.0013 against 0.0377$\pm$0.0004 for the best baseline), well beyond run-to-run variation. This margin reflects two distinct advantages. Our encoder assigns each modality its own pretrained patch-embedding adapter and concatenates the resulting tokens along the sequence axis, whereas the baselines can only accept the eight layers stacked as channels at a shared stem. The randomly initialized copy of our model isolates the two effects: it shares the tokenization scheme without lunar pretraining and already reaches RMSE 0.0397$\pm$ 0.0004, ahead of five of the seven ImageNet-pretrained baselines, while pretraining accounts for the remaining gap of 0.0293$\pm$0.0013. Our modality-count ablation also shows that the pretraining advantage for ice prospectivity is realized with as few as three input modalities (aspect, slope, and DICE), without requiring the full eight-modality stack. On NAC crater detection and IMP segmentation \lfm places among the leading entries, but the margins are smaller than the seed spread. Therefore, performance of \lfm and baselines can be seen as comparable.

A practical insight can be derived from the adaptation comparison performed in our experiments. LoRA is competitive with full fine-tuning throughout, and on crater detection it is the better choice outright: it tops every column of the WAC benchmark at full training data while also giving the best \lfm mAP on NAC. This is achieved by training a small fraction of the encoder parameters, which also leads to narrower seed spread than full fine-tuning. Full fine-tuning retains an edge only on the two smallest benchmarks, IMP segmentation and ice prospectivity, where the extra capacity to adapt appears to pay off. Freezing the encoder entirely is the least transferable option: it is the strongest configuration on IMP, where 100 training tiles make the reduced parameter count a useful regularizer, but falls below every baseline on crater detection. Based on our experiments, practitioners adapting \lfm to a new lunar task could treat LoRA as the sensible default, reserving full fine-tuning for small-sample regimes and validating a frozen encoder before relying on it.

Alongside quantitative fine-tuning results, the multimodal generation experiments confirm that pretraining has induced the cross-modal correspondences the objective was designed to capture: terrain derivatives from elevation, illumination-consistent reflectance from geometry, and coarse spatial pattern across the full co-registered modality set. These qualitative experiments also make the model's limits apparent: absolute elevation offsets and geodetic coordinates are not reliably recovered, as well as some static maps. This reinforces that the model is intended as a reusable representation for downstream tasks rather than a substitute for physical instruments.

Several limitations bound the present work. The controlled ablations needed to isolate the contribution of geometry tokenization and mixed-resolution training from lunar-domain pretraining as a whole remain to be run. Despite SomBench providing high quality labeled data, some of the benchmarks have limited samples in the test set. However, given the thorough process that went into creating such benchmarks, we believe they still represent an important validation instrument for lunar ML models. Finally, the generation experiments reveal that the model does not maintain a geodetic reference frame, a gap that is relevant to applications requiring absolute positioning. However, in principle, this could be learned during finetuning.

Future work will address these gaps through targeted ablations, extension to the full modality stack, and evaluation on additional planetary bodies where instrument constellations share structural similarities to the lunar case. We release the pretrained checkpoint, fine-tuning code, and dataset as open resources to support reproducibility and community reuse. 

\section*{Acknowledgments}
This work was supported by the National Aeronautics and Space Administration under Award No. 80MSFC25M0084.


\section*{Data and Code availability}
The data is available through SomBench \cite{sombench2026collection}. All the finetuning code to reproduce our results is available via \href{https://github.com/NASA-IMPACT/NASA-IBM-Lunar-Foundation-Model}{GitHub}. 

\section*{Appendix}

\subsection*{Tokenizer training details}

Table \ref{tab:tokenizer-training-config} shows details about the training configuration for each tokenizer.

\begin{table}[htpb]
\centering
\caption{Per-run training configuration for the nine domain-specific lunar tokenizers. All runs share a per-GPU batch size of 32.}
\label{tab:tokenizer-training-config}
\begin{tabular}{llccccc}
\toprule
Tokenizer & Modality & GPUs & Global Batch & Epochs & Warmup Epochs & Peak LR \\
\midrule
\multicolumn{7}{l}{\textit{NAC}} \\
\midrule
\texttt{nac\_aspect\_3m}  & Aspect             & 16 & 512 & 100 & 2 & 3.06e-6 \\
\texttt{nac\_dtm\_3m}     & DTM                & 16 & 512 & 100 & 2 & 3.06e-6 \\
\texttt{nac}              & Panchromatic       & 16 & 512 & 100 & 2 & 3.06e-6 \\
\texttt{nac\_slope\_3m}   & Slope              & 16 & 512 &  60 & 2 & 3.06e-6 \\
\midrule
\multicolumn{7}{l}{\textit{WAC}} \\
\midrule
\texttt{wac\_aspect}      & Aspect             &  8 & 256 & 100 & 1 & 1.57e-6 \\
\texttt{wac\_dtm}         & DTM                & 16 & 512 & 100 & 1 & 3.06e-6 \\
\texttt{wac\_slope}       & Slope              &  4 & 128 &  30 & 1 & 1.57e-6 \\
\texttt{wac\_uv}          & UV reflectance     &  4 & 128 &  50 & 1 & 1.57e-6 \\
\texttt{wac\_vis}         & Visible imagery    & 20 & 640 & 100 & 2 & 2.50e-5 \\
\bottomrule
\end{tabular}
\end{table}

\subsection*{Additional generation results}

\begin{figure}[ht]
    \centering
    \includegraphics[width=\linewidth, trim=0cm 0cm 0cm 0.9cm,
    clip]{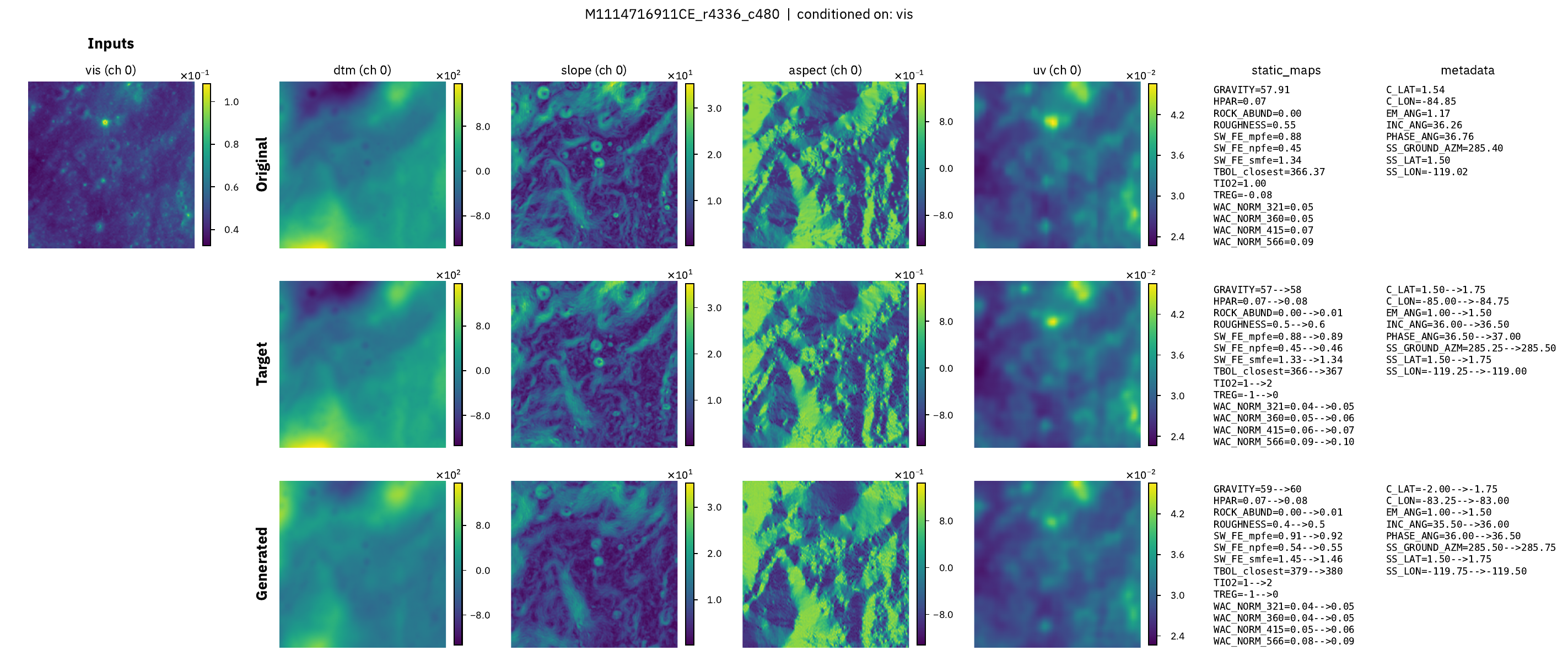}
    \vspace{0.5em}
    \includegraphics[width=\linewidth, trim=0cm 0cm 0cm 0.9cm,
    clip]{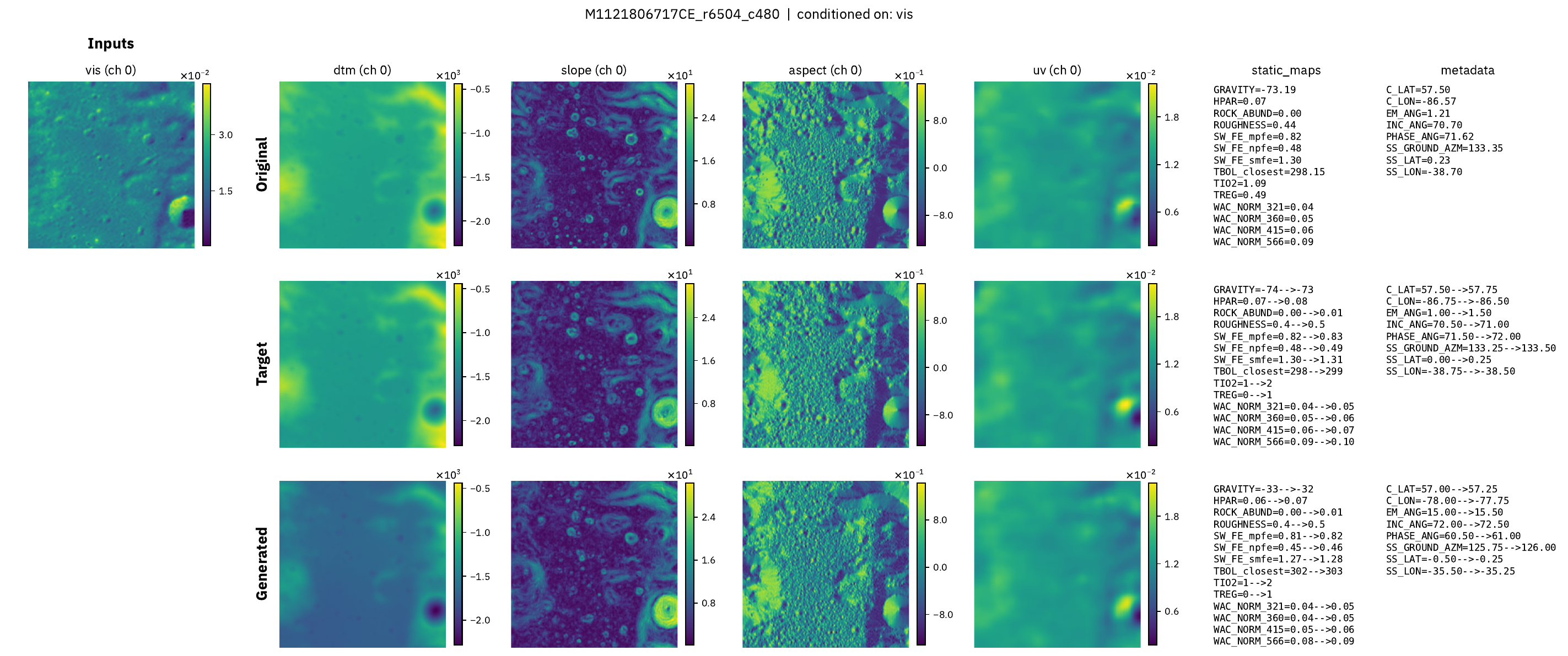}
    \caption{Additional generation examples from WAC to all other modalities.}
    \label{fig:vis2all_examples}
\end{figure}

\clearpage
\bibliographystyle{unsrt}  
\bibliography{references}

\end{document}